\documentclass{article} 
\usepackage{redis_format,times}
\usepackage{arxiv_preprint}

\usepackage{amsmath,amsfonts,bm}

\def\eqref#1{equation~\ref{#1}}

\def\1{\bm{1}}

\DeclareMathAlphabet{\mathsfit}{\encodingdefault}{\sfdefault}{m}{sl}
\SetMathAlphabet{\mathsfit}{bold}{\encodingdefault}{\sfdefault}{bx}{n}

\usepackage{hyperref}
\usepackage{url}
\usepackage{xcolor}
\usepackage{graphicx}
\usepackage{booktabs}
\usepackage{wrapfig}
\usepackage{needspace}
\usepackage{capt-of}
\usepackage{tcolorbox}
\tcbuselibrary{breakable,skins}
\usepackage{fvextra}
\newtcolorbox{promptbox}[2][]{%
    enhanced, breakable,
    colback=white, colframe=black!65, colbacktitle=black!65,
    coltitle=white, fonttitle=\bfseries\small, fontupper=\small,
    title={#2}, title after break={#2 (continued)},
    boxrule=0.5pt, arc=1mm, left=3mm, right=3mm,
    top=2mm, bottom=2mm, toptitle=1mm, bottomtitle=1mm,
    before skip=7pt, after skip=7pt,
    before upper={\raggedright\setlength{\parindent}{0pt}\setlength{\parskip}{2pt}},
    #1
}
\usepackage{amsmath}
\usepackage{amssymb}
\usepackage{booktabs}
\usepackage{multirow}
\usepackage{xcolor}
\usepackage{graphicx}
\usepackage{enumitem}

\definecolor{lycolor}{RGB}{0, 0, 0}
\newcommand{\ly}[1]{{\color{lycolor}#1}}

\definecolor{wtcolor}{RGB}{0, 0, 0}
\newcommand{\wt}[1]{{\color{wtcolor}#1}}

\definecolor{deltacolor}{RGB}{50,110,170}
\usepackage{booktabs,multirow,graphicx,xcolor}
\usepackage{wrapfig,caption,adjustbox,ragged2e,needspace}

\newsavebox{\ReDiSAblationTableBox}
\newsavebox{\ReDiSAblationFigureBox}

\usepackage{caption}
\usepackage{adjustbox}
\usepackage{needspace}
\hypersetup{hidelinks}
\title{ReDiS: Reasoning–Diagnosis Self-Evolution for Multimodal Reasoning in Remote Sensing}

\newcommand{\paperauthors}{%
  Yi Liu$^{1,2}$, Wentao Jiang$^{1}$, Shuo Ni$^{3,2}$, Zhiming Luo$^{1,2}$,   Di Wang$^{1,2}$, Zengmao Wang$^{1}$, \\ 
  Jing Zhang$^{1,2,\dagger}$, Bo Du$^{1,2,\dagger}$}
\newcommand{\paperaffiliations}{%
  $^{1}$School of Computer Science, Wuhan University\quad 
  $^{2}$Zhongguancun Academy\quad \\
  $^{3}$National Key Laboratory of Science and Technology on Space-Born Intelligent Information Processing, Beijing Institute of Technology}
\definecolor{emailblue}{RGB}{62,132,194}
\newcommand{\paperemails}{%
  \href{mailto:jingzhang.cv@gmail.com}{jingzhang.cv@gmail.com}{\color{red};}\quad
  \href{mailto:dubo@whu.edu.cn}{dubo@whu.edu.cn}}
\author{\paperauthors}
\date{}
\hypersetup{
  pdftitle={ReDiS: Reasoning–Diagnosis Self-Evolution for Multimodal Reasoning in Remote Sensing},
  pdfauthor={Yi Liu, Wentao Jiang, Shuo Ni, Zhiming Luo, Di Wang, Zengmao Wang, Jing Zhang, Bo Du},
  pdfsubject={ReDiS: Reasoning–Diagnosis Self-Evolution for Multimodal Reasoning in Remote Sensing},
  pdfkeywords={ReDiS, remote sensing, visual question answering, hallucination}
}

\begin{document}

\maketitle

\begin{abstract}

Multimodal large language models (MLLMs) have shown strong potential for remote sensing understanding, yet reliable reasoning remains challenging when tasks require fine-grained visual grounding and multi-step inference. Existing approaches typically rely on either costly pre-constructed reasoning trajectories or self-generated signals such as response consistency, which can reinforce shared errors and provide unreliable supervision. We introduce \textbf{Reasoning--Diagnosis Self-Evolution (ReDiS)}, a framework that separates reasoning exploration from candidate evaluation while allowing the two processes to evolve jointly. ReDiS instantiates two role-specific experts from a shared multimodal backbone: a \textbf{Reasoner} that explores diverse visual evidence--answer candidates and a \textbf{Diagnoser} that jointly evaluates the complete candidate set and provides credibility-aware feedback. As the Reasoner evolves, it exposes new candidate distributions and failure modes; the Diagnoser adapts to these changes and, in turn, reshapes the learning signal for subsequent exploration. To handle groups in which all sampled answers are incorrect and group-relative advantages vanish, we further introduce \textbf{Failure-Triggered Recovery (FTR)}, which provides minimal outcome-level correction without prescribing intermediate reasoning trajectories. Experiments on LHRS-Bench, MME-RS, and CHOICE show consistent improvements over Qwen3.5 backbones at 0.8B, 2B, and 4B scales. ReDiS improves category-averaged accuracy by 6.47 percentage points at 0.8B, while ReDiS-4B achieves the highest average performance among the compared remote sensing MLLMs. Code and data will be publicly available at
\href{https://github.com/MiliLab/ReDiS}{\texttt{https://github.com/MiliLab/ReDiS}}.
\end{abstract}

\begin{figure}[h]
    \centering
    \includegraphics[width=0.93\linewidth]{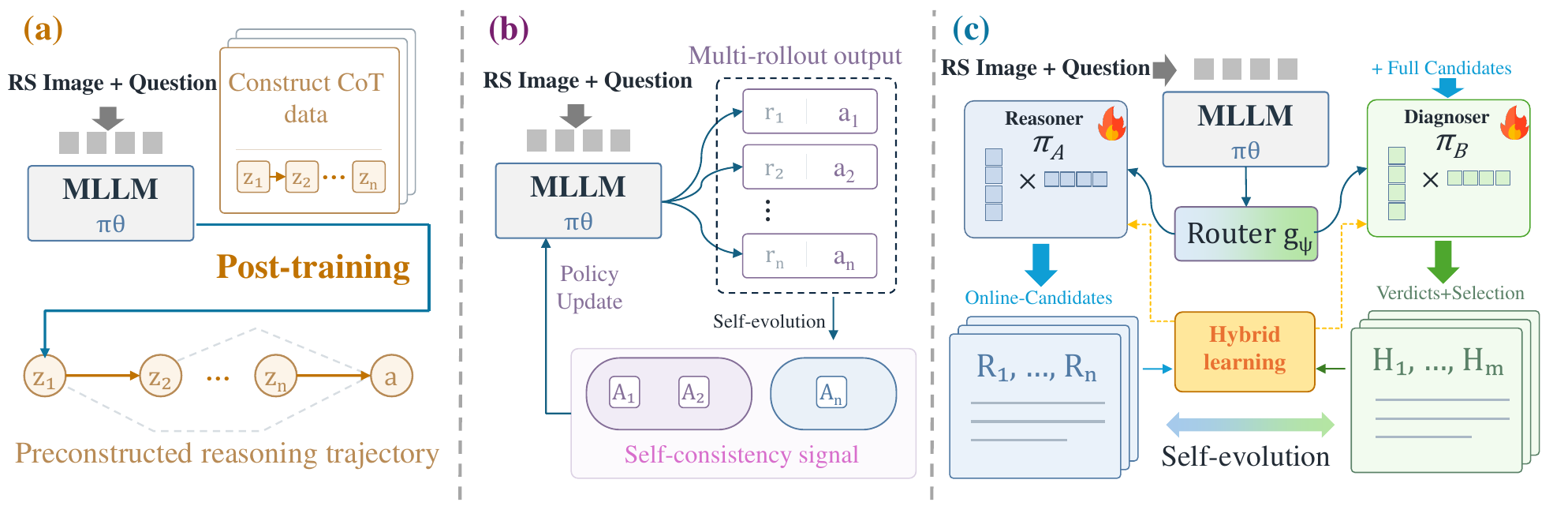}
    \caption{Training paradigms for remote sensing reasoning. (a) Trajectory-supervised methods learn from pre-constructed chains of thought. (b) Self-consistency methods derive feedback from multiple online rollouts. (c) The Reasoning–Diagnosis Self-Evolution (ReDiS) framework for multimodal reasoning instantiates two role-specific experts from a shared multimodal model.}
    \label{fig:motivation}
\end{figure}

\section{Introduction}

\ly{
MLLMs ~\citep{yang2025qwen3,dai2023instructblip,chen2024open,liu2023visual} have advanced remote sensing interpretation, from image captioning ~\citep{cheng2022nwpu,wang2026dppf} and visual question answering ~\citep{lobry2020rsvqa,wang2024earthvqa,chappuis2022prompt} to visual grounding and geospatial reasoning ~\citep{yao2026remotereasoner,liu2026towards}. However, reliable reasoning remains challenging in scenes with small, densely distributed objects and complex spatial relationships~\citep{yu2006object}. Answering questions in these scenes requires identifying subtle visual evidence and integrating multiple observations. Errors in recognizing objects, attributes, or relations can propagate through subsequent inference, producing plausible answers based on unsupported visual premises. Reliable remote sensing interpretation therefore requires both correct answers and reasoning grounded in question-relevant visual evidence.
}

\ly{
This exposes the limitation of simple answer-level supervision: it can verify whether a final prediction is correct, but provides little information about the reasoning process. A common approach is to construct reasoning trajectories and use them as intermediate supervision~\citep{yao2026remotereasoner}. Such trajectories provide richer guidance, but high-quality traces are expensive to generate and filter, and they are fixed before policy optimization. As the policy evolves, its errors and hallucination patterns may no longer be well covered by these pre-constructed traces. Self-evolution methods obtain feedback from online rollouts of the current policy~\citep{thawakar2025evolmm}. Although these rollouts reflect evolving model behavior, agreement among them need not indicate visual correctness: multiple responses may reinforce the same visual misconception~\citep{venkatraman2026paying}. Recent methods strengthen visual feedback through invariance constraints~\citep{venkatraman2026paying} or region-level verification~\citep{yao2026remotezero}. These advances motivate a complementary question: \textbf{\textit{how can reasoning exploration and joint candidate diagnosis evolve together while keeping self-generated feedback anchored to task correctness?}}
}


\wt{To address this, we take a different design, which separates exploration from evaluation while allowing them to evolve together. One process explores alternative visual evidence and answers, while the other compares the resulting candidates and judges their reliability. As exploration evolves, it continually changes the candidate distribution and failure modes that evaluation must learn to distinguish. The evaluator, in turn, changes the learning signal seen by exploration, reshaping which behaviors are reinforced in the next round. Their evolution remains constrained by answer-level supervision, which fixes the criterion of correctness outside this loop.}

\wt{We introduce ReDiS, a Reasoning--Diagnosis Self-evolution framework built around two role-specific experts: a Reasoner and a Diagnoser. Figure~\ref{fig:motivation} contrasts ReDiS with trajectory-supervised and self-consistency paradigms. Given an image--question pair, the Reasoner samples a set of evidence--answer candidates from its current policy. The Diagnoser observes the image, question, and the complete candidate set, assigns a diagnostic state to each candidate, and either selects a preferred response or rejects the entire set. After the rollouts are generated, the reference answer is used to construct correctness labels for the Reasoner and answer-anchored targets for the Diagnoser; both roles are then optimized with group-relative objectives. The Diagnoser's judgments are further aggregated into candidate-level credibility scores, which refine the Reasoner's reward only for answer-correct candidates, while incorrect candidates receive no positive reward regardless of the diagnosis.
}
\wt{
A remaining failure case arises when exploration produces no correct candidate.
In this case, correctness-gated rewards collapse across the group, leaving group-relative optimization with no direction to improve. We address this with Failure-Triggered Recovery (FTR), which intervenes only on failed groups and supplies minimal supervision to the answer value tokens. FTR therefore complements the reasoning--diagnosis loop: diagnosis guides learning when exploration is informative, while minimal correction is used only when exploration provides no usable signal.
}

Experiments on three remote sensing visual question answering (RSVQA) benchmarks, including LHRS-Bench, MME-RS, and CHOICE, show consistent improvements over the Qwen3.5-0.8B, 2B, and 4B backbones. At 0.8B, ReDiS improves average performance by 6.47 percentage points; at 4B, it achieves the best overall performance among the compared models. Ablation studies further support the contributions of diagnostic feedback and FTR. 

Our main contributions are threefold:
\begin{itemize}[
    leftmargin=*,
    labelindent=0pt,
    labelsep=0.5em,
    topsep=4pt,
    itemsep=3pt,
    parsep=0pt,
    partopsep=0pt
]
    \item We introduce ReDiS, which jointly evolves reasoning exploration and diagnosis for RSVQA. Answer-anchored diagnostic targets and correctness-gated feedback connect the two roles using answer labels without annotated reasoning trajectories.
    \item We propose FTR to address the loss of relative learning signals in all-incorrect Reasoner groups. Minimal outcome-level correction provides an optimization direction without prescribing visual evidence or reasoning targets.
    \item Experiments on LHRS-Bench, MME-RS, and CHOICE demonstrate improvements over the Qwen3.5-0.8B, 2B, and 4B backbones. ReDiS improves AA by 6.47 percentage points on average at 0.8B, while ReDiS-4B achieves the highest average OA and AA among the compared models, including larger remote sensing MLLMs.
\end{itemize}

\section{Related Work}
\subsection{Multimodal Large Language Models in Remote Sensing}
With the development of general MLLMs \citep{yang2025qwen3, dai2023instructblip,chen2024open,liu2023visual}, remote sensing research is gradually shifting from task-specific models toward unified multimodal modeling frameworks. RSGPT \citep{hu2025rsgpt} investigates the transferability of large vision-language models to remote sensing image captioning and RSVQA through high-quality remote sensing image-text data. GeoChat \citep{kuckreja2024geochat} further extends these capabilities to region-level dialogue, visual grounding, and high-resolution remote sensing understanding. EarthGPT \citep{zhang2024earthgpt} integrates multi-source data, including optical, SAR, and infrared imagery, and supports a diverse set of tasks, such as scene classification, visual question answering, grounding, and object detection, within a unified model. Subsequent studies further advance domain-specific visual understanding. LHRS-Bot \citep{muhtar2024lhrs} improves remote sensing understanding through large-scale image-text data and multi-level vision-language alignment, while GeoLLaVA-8K \citep{wang2026geollava} addresses visual token redundancy in ultra-high-resolution remote sensing imagery through background token pruning and anchored token selection. 

Beyond advances in visual perception, recent studies have also begun to investigate explicit reasoning in remote sensing models. RSThinker \citep{liu2026towards} constructs perceptually-grounded geospatial chain-of-thought and applies supervised fine-tuning with reinforcement learning to improve answer correctness. These methods demonstrate the effectiveness of process-level supervision for complex remote sensing reasoning, but typically require additional construction and filtering of high-quality reasoning trajectories. Rather than providing predefined target reasoning trajectories, ReDiS allows the current policy to generate candidates online and shift the optimization objective from reasoning trajectory imitation to online reasoning exploration and diagnosis.

\subsection{Self-Evolution and Internal Feedback Learning}
Self-evolution has recently emerged as an important direction for improving reasoning models. In language models, EMPO \citep{zhang2026right} constructs unsupervised reward signals from semantic entropy and improves reasoning by reinforcing high-frequency semantic clusters. INTUITOR~\citep{zhao2026learning} proposes Reinforcement Learning from Internal Feedback (RLIF), which replaces external correctness rewards with internal self-certainty and incorporates this signal into group-relative optimization with GRPO. This paradigm has subsequently been extended to multimodal self-evolution. Seir\^{e}nes~\citep{zhang2026seirenes} uses adversarial self-play to co-evolve distraction generation and mathematical problem solving within a shared model. EvoLMM \citep{thawakar2025evolmm} instantiates a Proposer and a Solver from the same model and improves visual reasoning through online question generation and answer self-consistency using only unlabeled images.

However, internal certainty does not necessarily indicate task correctness. No Free Lunch \citep{zhang2025no} systematically investigates internal feedback signals, and shows that they may induce excessive policy confidence and subsequent performance degradation during late training. Self-Judge~\citep{wu2026models} further observes that majority responses can converge to an incorrect consensus and therefore introduces a frozen Judge to regulate the quality of self-consistency signals. VISE \citep{venkatraman2026paying} characterizes a related failure mode as visual under-conditioning and introduces geometric and semantic invariance rewards to directly strengthen the dependence of the model on visual content. DUEL~\citep{qiu2026duel} instead constructs closely matched true and false statements through adversarial self-play between a Challenger and a Solver, requiring the Solver to perform fine-grained verification against the image. Collectively, these studies reflect a broader shift from answer-level self-consistency toward explicit visual verification. Nevertheless, most existing methods derive feedback from the same policy, a frozen auxiliary judge, or task-specific verification signals. 
In contrast, ReDiS moves beyond self-contained internal feedback by separating reasoning exploration from diagnosis and coupling their evolution through externally anchored supervision.
\begin{figure}[t]
    \centering
    \includegraphics[width=\linewidth]{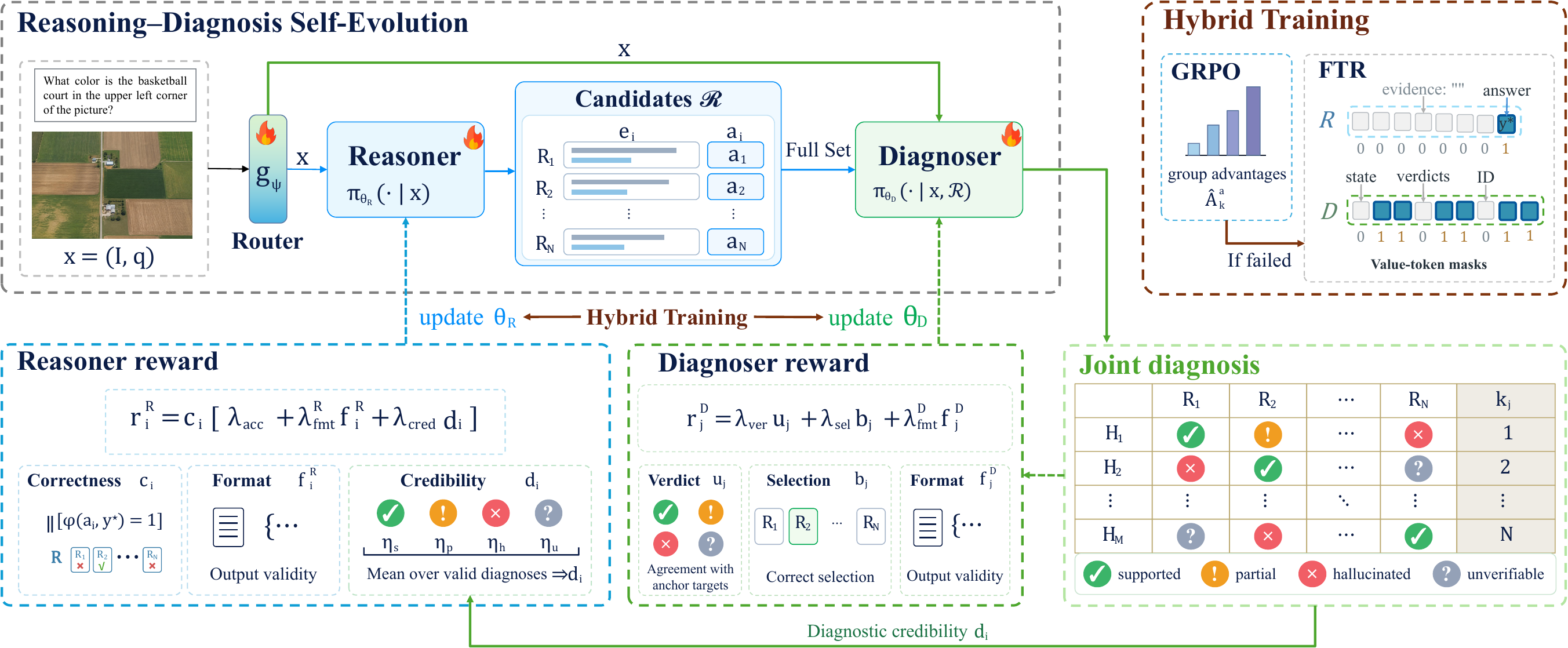}
    \caption{\textbf{Overview of ReDiS: Reasoning--Diagnosis Self-Evolution.} The Reasoner explores visual evidence--answer candidates, while the Diagnoser jointly evaluates the complete set. Answer labels anchor diagnostic learning and gate credibility feedback to correct-answer candidates. The two roles co-evolve through online sampling and separate group-relative updates. Failure-Triggered Recovery (FTR) supplies value-token correction on failed groups, leaving visual evidence and free-form reasoning without prescribed targets.}
    \label{fig:net1}
\end{figure}

\section{Method}
\subsection{Design Overview}
ReDiS learns directly from the model's evolving reasoning attempts while keeping self-generated feedback anchored to task correctness. A correct final answer alone does not establish whether its supporting visual claims are reliable, while agreement among multiple responses may simply reflect a shared error. We therefore decouple candidate generation from candidate evaluation. The \textbf{Reasoner} explores alternative visual evidence and answers, whereas the \textbf{Diagnoser} evaluates these candidates against the image and question. Reference answers are introduced only after rollout to anchor optimization, ensuring that diagnostic preferences cannot convert an incorrect candidate into a positive learning signal. Figure~\ref{fig:net1} illustrates this learning loop.

For each training example, the input $x=(I,q)$ consists of an image and a question, with reference answer $y^\star$. Both roles share a frozen multimodal backbone and use independent LoRA adapters~\citep{hu2022lora}, parameterized by $\theta_R$ and $\theta_D$, allowing the two roles to specialize without duplicating the full model. A separately trained router $g_\psi$ selects the corresponding role according to the input prompt. Neither role observes $y^\star$ during rollout; the reference answer is used only afterward to construct rewards and, when necessary, recovery targets. Training therefore requires neither annotated visual evidence nor teacher-generated reasoning trajectories. At inference time, only the Reasoner branch is activated for a single deterministic generation.

\subsection{Online Exploration and Joint Diagnosis}
\textbf{Exploring current reasoning attempts.}
Given $x$, the Reasoner samples multiple responses to expose alternative interpretations of the same image. Each candidate pairs a concise description of visual evidence or reasoning rationale $e_i$ with an answer $a_i$:
\begin{equation}
R_i=(e_i,a_i)\sim\pi_{\theta_R}(\cdot\mid x),
\qquad i=1,\ldots,N.
\label{eq:redis_candidates}
\end{equation}
The resulting candidates form the set $\mathcal{R}$. Because they are generated online, they reflect the current policy's visual interpretations and failure modes and therefore evolve together with the Reasoner. The explicit evidence field exposes the stated basis of each prediction to the Diagnoser without requiring a reference reasoning trajectory.

\textbf{Diagnosing the complete candidate set.}
The Diagnoser receives the input $x$ together with all $N$ candidates. Observing the complete candidate set allows it to compare the candidates globally and determine which response is the most reliable. Importantly, this global comparison does not make candidate-level judgments dependent on agreement among responses: each candidate is evaluated independently according to whether its answer and stated evidence are supported by the image and question. The Diagnoser samples $M$ complete diagnostic responses. Each sampled diagnosis assigns a verdict $\hat{v}_{ji}$ to every candidate and returns a selection $k_j$, where $k_j\in\{1,\ldots,N\}$ identifies the preferred candidate and $k_j=0$ indicates rejection of the entire set.

We use four diagnostic states: $v_s$ for visually supported content, $v_p$ for a supported answer accompanied by locally unreliable evidence, $v_h$ for an evident hallucination, and $v_u$ when the available evidence is insufficient for a reliable judgment. Candidate-level verdicts provide fine-grained feedback for individual reasoning candidates, whereas set-level selection identifies the most reliable candidate, with the option to reject the entire set. These visual judgments are learned through answer-anchored proxy targets, described next.

\subsection{Hybrid Training}
\textbf{Reasoner Reward.} After rollout, we determine candidate correctness by matching $a_i$ to the reference answer $y^\star$ with the answer-matching function $\phi$. This yields a binary indicator $c_i$, where unmatched or unparseable answers are treated as incorrect. We then aggregate valid diagnostic judgments into a candidate-level credibility score.
\begin{equation}
d_i=\frac{1}{|\mathcal{J}_{\mathrm{valid}}|}
\sum_{j\in\mathcal{J}_{\mathrm{valid}}}s(\hat v_{ji}),
\label{eq:redis_credibility}
\end{equation}
Here, $\mathcal{J}_{\mathrm{valid}}$ denotes the set of diagnostic responses satisfying the candidate-binding constraints, and $d_i=0$ when no valid diagnosis is available. The scoring function $s(\cdot)$ maps the four verdicts to the nonnegative values $\eta_s$, $\eta_p$, $\eta_h$, and $\eta_u$. We combine this credibility signal with answer correctness and structured-output validity to define the Reasoner reward:
\begin{equation}
r_i^R=c_i\left(\lambda_{\mathrm{acc}}
+\lambda_{\mathrm{fmt}}^Rf_i^R+\lambda_{\mathrm{cred}}d_i\right).
\label{eq:redis_reasoner_reward}
\end{equation}
Here, the nonnegative weights $\lambda_{\mathrm{acc}}$, $\lambda_{\mathrm{fmt}}^R$, and $\lambda_{\mathrm{cred}}$ control correctness, format, and credibility, respectively. The multiplicative correctness gate is crucial: every incorrect candidate receives zero reward regardless of how strongly the Diagnoser endorses it. Diagnostic feedback therefore refines preferences only among answer-correct candidates, providing an additional learning signal beyond the shared outcome label without allowing an incorrect consensus to receive positive reinforcement.

\textbf{Diagnoser Reward.}
Because the Diagnoser itself evolves, its predictions require an external anchor during optimization. We derive proxy targets for the diagnostic states defined above from answer correctness: a correct answer maps to $v_s$, a parseable incorrect answer to $v_h$, and an unparseable answer to $v_u$. The state $v_p$ cannot be inferred directly from the answer label, since a correct answer may still contain locally unreliable evidence. We therefore treat $v_p$ as a softened match to $v_s$: when the proxy target is $v_s$, predicting $v_p$ receives partial credit $\kappa\in[0,1]$ rather than being counted as a complete mismatch. The verdict-consistency score $u_j$ is then computed by averaging these per-candidate matching scores over all $N$ candidates.

We reward set-level selection separately. If at least one candidate is answer-correct, the selection score $b_j$ equals one only when the Diagnoser selects a correct candidate whose predicted verdict is either $v_s$ or $v_p$. If no candidate is correct, $b_j=1$ only when the complete set is rejected, i.e., $k_j=0$. Otherwise, $b_j=0$. Combining verdict consistency, selection quality, and structured-output validity yields the Diagnoser reward:
\begin{equation}
r_j^D=\lambda_{\mathrm{ver}}u_j
+\lambda_{\mathrm{sel}}b_j+\lambda_{\mathrm{fmt}}^Df_j^D,
\label{eq:redis_diagnoser_reward}
\end{equation}
The nonnegative weights $\lambda_{\mathrm{ver}}$, $\lambda_{\mathrm{sel}}$, and $\lambda_{\mathrm{fmt}}^D$ control candidate-level diagnostic consistency, set-level decision quality, and output validity, respectively. These answer-derived targets serve only as outcome anchors; they do not directly supervise evidence faithfulness or diagnostic rationales.

\textbf{Self-Evolution Framework.} The two roles interact through a changing online learning distribution. As the Reasoner improves, it produces new candidate sets and exposes different error patterns to the Diagnoser. As the Diagnoser evolves, its credibility estimates reshape the relative preference among successful Reasoner candidates. Repeated online sampling closes this loop: each role continually changes the training distribution or feedback observed by the other, while reference answers remain an external anchor for task correctness. Each role performs group-relative optimization within its own rollout group. The Reasoner compares $N$ candidates generated for the same input, whereas the Diagnoser compares $M$ complete diagnostic responses for the same candidate set. For role $a\in\{R,D\}$, group rewards are normalized as
\begin{equation}
\hat A_k^a=\frac{r_k^a-\mu_a}{\sigma_a+\epsilon},
\qquad G_R=N,\quad G_D=M,
\label{eq:redis_advantage}
\end{equation}
where $G_a$ denotes the corresponding group size, and $\mu_a$ and $\sigma_a$ are the group reward mean and standard deviation. The constant $\epsilon>0$ ensures numerical stability. Using separate optimizers, the two role-specific adapters independently maximize the clipped GRPO objective:
\begin{equation}
\mathcal{J}_{\mathrm{GRPO}}^a=\mathbb{E}\left[
\frac{1}{G_a}\sum_{k=1}^{G_a}\frac{1}{T_k}\sum_{t=1}^{T_k}
\min\left(
\rho_{k,t}^a(\theta_a)\hat A_k^a,
\operatorname{clip}\left(\rho_{k,t}^a(\theta_a),1-\varepsilon,1+\varepsilon\right)\hat A_k^a
\right)\right].
\label{eq:redis_grpo}
\end{equation}
Here, $T_k$ denotes the rollout length, $\rho^a_{k,t}$ is the current-to-rollout policy probability ratio for token $t$ under the same input and prefix, and $\varepsilon$ is the clipping threshold. The expectation is taken over training inputs and online rollouts.

\textbf{Failure-Triggered Recovery.}\label{sec:ftr} Correctness gating prevents incorrect candidates from receiving positive reward, but it also exposes a degenerate case of group-relative optimization. When all $N$ Reasoner candidates are incorrect, every Reasoner reward becomes zero and the corresponding group-relative advantages vanish. No sampled response then provides a useful direction for GRPO. We introduce \textbf{Failure-Triggered Recovery (FTR)} to restore a learning signal specifically for these failed groups while leaving intermediate reasoning unconstrained. 

For the Reasoner, the failure set $\mathcal{F}_R$ contains inputs for which all $N$ sampled candidates are incorrect. Recovery supervises only the reference-answer value tokens and provides neither target visual evidence nor a target reasoning trajectory. For the Diagnoser, $\mathcal{F}_D$ contains inputs for which none of the $M$ diagnostic responses simultaneously provides valid candidate-level judgments and a valid set-level decision. Its recovery targets follow the answer-anchored verdict and selection rules defined above, with supervision restricted to diagnostic-state and candidate-selection value tokens. Free-form diagnostic rationales remain unsupervised. 

We implement this restriction through a binary value-token mask $m^a_{x,t}$ over the correction target $z_x^a$. For a nonempty failure set, the masked cross-entropy objective is
\begin{equation}
\mathcal{L}_{\mathrm{corr}}^a=
-\frac{1}{|\mathcal{F}_a|}\sum_{x\in\mathcal{F}_a}
\frac{\sum_t m_{x,t}^a\log\pi_{\theta_a}
\left(z_{x,t}^a\mid\xi_x^a,z_{x,<t}^a\right)}
{\sum_t m_{x,t}^a},
\label{eq:redis_recovery}
\end{equation}
where the conditioning input $\xi_x^a$ is $x$ for the Reasoner and $(x,\mathcal{R})$ for the Diagnoser. For each role, the correction step is applied after the GRPO update and is skipped when the corresponding failure set is empty. GRPO and FTR therefore remain separate optimizer steps rather than components of a weighted joint loss. Group-relative learning exploits differences among sampled responses whenever informative candidates are available, whereas FTR supplies a minimal outcome-directed signal only when those relative differences disappear.

\section{Experimental Setup}
\textbf{Training setup.} We implement ReDiS in VERL \citep{verl} and instantiate the Reasoner and Diagnoser as two independent rank-16 LoRA~\citep{hu2022lora} adapters over a frozen Qwen3.5 multimodal backbone. Training uses the GeoLLaVA-8K ~\citep{wang2026geollava} data release with answer labels only; no reference visual evidence or teacher-generated reasoning trajectories are used. For each question, the Reasoner samples $N=8$ candidates and the Diagnoser produces $M=8$ diagnostic responses. Both adapters are optimized with AdamW~\citep{loshchilov2019decoupled} using a base learning rate of $10^{-6}$ for up to 1,000 training steps. We set $(\lambda_{\mathrm{acc}},\lambda_{\mathrm{fmt}}^R,\lambda_{\mathrm{cred}})=(2,0.1,0.25)$ and $(\lambda_{\mathrm{ver}},\lambda_{\mathrm{sel}},\lambda_{\mathrm{fmt}}^D)=(2,0.5,0.5)$. The verdict scores are $(\eta_s,\eta_p,\eta_h,\eta_u)=(1,0.5,0,0)$, with partial-credit coefficient $\kappa=0.5$. Additional implementation, sampling, and optimization details are provided in Appendix~\ref{app:experimental_details}.

\paragraph{Benchmarks and Baselines.} We evaluate RSVQA on LHRS-Bench~\citep{muhtar2024lhrs}, MME-RS from MME-RealWorld~\citep{zhang2024mmerealworld}, and CHOICE~\citep{an2025choice}. We report overall accuracy (OA) and category-averaged accuracy (AA); Avg. denotes the unweighted mean across benchmarks. Table~\ref{tab:main_results} compares ReDiS with three groups of MLLMs: proprietary models (GPT-4o \citep{hurst2024gpt}, Gemini-2.5-Pro \citep{comanici2025gemini}, and Claude-Haiku-4.5); open-source general-purpose models (Qwen3.5 \citep{qwen35blog}, LLaVA-v1.6-Vicuna \citep{liu2023visual}, LLaMA3-LLaVA \citep{chen2024open}, and DeepSeek-VL-Chat \citep{lu2024deepseekvl}); and remote sensing models (GeoChat \citep{kuckreja2024geochat}, VHM \citep{pang2025vhm}, EarthDial \citep{soni2025earthdial}, RemoteReasoner \citep{yao2026remotereasoner}, and RSThinker \citep{liu2026towards}). Qwen3.5 at 0.8B, 2B, and 4B provides the corresponding backbone references; each $\Delta$ reports the absolute gain in percentage points over the backbone at the same scale.

\section{Results and Analysis}
\subsection{Quantitative Analysis}
\begin{table*}[t]
\centering
\caption{Comparison with existing vision-language models on remote sensing benchmarks. OA and AA denote overall accuracy and average accuracy, respectively.}
\label{tab:main_results}
\setlength{\tabcolsep}{5.0pt}
\renewcommand{\arraystretch}{1.08}
\small
\resizebox{\textwidth}{!}{
\begin{tabular}{lcccccccccc}
\toprule
\multirow{2}{*}{\textbf{Method}} & \multirow{2}{*}{\textbf{Publication}} & \multirow{2}{*}{\textbf{Size}} & \multicolumn{2}{c}{\textbf{LHRS-Bench}} & \multicolumn{2}{c}{\textbf{MME-RS}} & \multicolumn{2}{c}{\textbf{CHOICE}} & \multicolumn{2}{c}{\textbf{Avg.}} \\
\cmidrule(lr){4-5}\cmidrule(lr){6-7}\cmidrule(lr){8-9}\cmidrule(lr){10-11}
& & & \textbf{OA} & \textbf{AA} & \textbf{OA} & \textbf{AA} & \textbf{OA} & \textbf{AA} & \textbf{OA} & \textbf{AA} \\
\midrule

\multicolumn{11}{l}{\textit{\textcolor{gray}{Closed-source Commercial Vision-Language Models}}} \\
GPT-4o & arXiv 2024 & -- & 70.29 & 66.15 & 21.43 & 21.36 & 67.71 & 65.66 & 53.14 & 51.06 \\
Gemini-2.5-pro & arXiv 2025 & -- & \underline{74.35} & 66.41 & 38.23 & 38.06 & \underline{75.84} & \underline{72.48} & \underline{62.81} & 58.98 \\
Claude-haiku-4.5 & -- & -- & 58.70 & 56.59 & 20.48 & 21.03 & 64.32 & 59.82 & 47.83 & 45.81 \\

\midrule
\multicolumn{11}{l}{\textit{\textcolor{gray}{Open-source Vision-Language Models}}} \\
Qwen3.5-0.8B & -- & 0.8B & 54.35 & 45.40 & 38.36 & 38.36 & 56.79 & 51.90 & 49.83 & 45.22 \\
Qwen3.5-2B & -- & 2B & 65.80 & 56.24 & 36.49 & 36.34 & 63.74 & 62.54 & 55.34 & 51.71 \\
Qwen3.5-4B & -- & 4B & 72.20 & \underline{69.54} & 38.63 & 38.63 & 72.16 & 71.68 & 61.00 & \underline{59.95} \\
LLaVA-v1.6-vicuna & NeurIPS 2023 & 7B & 60.72 & 52.87 & 38.98 & 38.87 & 56.29 & 53.32 & 52.00 & 48.35 \\
LLaMA3-LLaVA & -- & 8B & 66.09 & 55.50 & \textbf{47.57} & \textbf{47.57} & 65.99 & 63.47 & 59.88 & 55.51 \\
DeepSeek-VL-Chat & arXiv 2024 & 7B & 57.10 & 47.98 & 35.07 & 34.96 & 63.37 & 62.07 & 51.85 & 48.34 \\

\midrule
\multicolumn{11}{l}{\textit{\textcolor{gray}{Open-source Remote Sensing Vision-Language Models}}} \\
GeoChat & CVPR 2024 & 7B & 49.86 & 44.09 & 29.29 & 29.24 & 55.11 & 52.89 & 44.75 & 42.07 \\
VHM & AAAI 2025 & 7B & 54.78 & 49.42 & 38.47 & 38.38 & 63.13 & 58.35 & 52.13 & 48.72 \\
EarthDial & CVPR 2025 & 4B & 56.81 & 49.30 & 33.98 & 33.88 & 63.22 & 59.75 & 51.34 & 47.64 \\
RemoteReasoner & AAAI 2026 & 7B & 66.67 & 57.18 & 30.44 & 30.29 & 71.29 & 69.02 & 56.13 & 52.16 \\
RSThinker & ICLR 2026 & 9B & 65.51 & 58.16 & 38.44 & 38.36 & 62.33 & 56.70 & 55.43 & 51.07 \\

\midrule
\multicolumn{11}{l}{\textit{\textcolor{gray}{Our Method}}} \\
ReDiS-0.8B & \multirow{2}{*}{Ours} & \multirow{2}{*}{0.8B} & 57.83 & 49.87 & \underline{46.28} & \underline{46.15} & 63.13 & 59.05 & 55.75 & 51.69 \\
\textcolor{deltacolor}{$\Delta$} & & & \textcolor{deltacolor}{+3.48} & \textcolor{deltacolor}{+4.47} & \textcolor{deltacolor}{+7.92} & \textcolor{deltacolor}{+7.79} & \textcolor{deltacolor}{+6.34} & \textcolor{deltacolor}{+7.15} & \textcolor{deltacolor}{+5.91} & \textcolor{deltacolor}{+6.47} \\

ReDiS-2B & \multirow{2}{*}{Ours} & \multirow{2}{*}{2B} & 70.14 & 59.97 & 44.44 & 44.59 & 71.84 & 69.65 & 62.14 & 58.07 \\
\textcolor{deltacolor}{$\Delta$} & & & \textcolor{deltacolor}{+4.34} & \textcolor{deltacolor}{+3.73} & \textcolor{deltacolor}{+7.95} & \textcolor{deltacolor}{+8.25} & \textcolor{deltacolor}{+8.10} & \textcolor{deltacolor}{+7.11} & \textcolor{deltacolor}{+6.80} & \textcolor{deltacolor}{+6.36} \\

\textbf{ReDiS-4B} & \multirow{2}{*}{Ours} & \multirow{2}{*}{4B} & \textbf{74.49} & \textbf{72.42} & 39.54 & 39.35 & \textbf{77.23} & \textbf{75.03} & \textbf{63.75} & \textbf{62.27} \\
\textcolor{deltacolor}{$\Delta$} & & & \textcolor{deltacolor}{+2.29} & \textcolor{deltacolor}{+2.88} & \textcolor{deltacolor}{+0.91} & \textcolor{deltacolor}{+0.72} & \textcolor{deltacolor}{+5.07} & \textcolor{deltacolor}{+3.35} & \textcolor{deltacolor}{+2.76} & \textcolor{deltacolor}{+2.32} \\
\bottomrule
\end{tabular}}
\end{table*}

Table~\ref{tab:main_results} shows that ReDiS consistently improves its matched Qwen3.5 backbone across all three benchmarks and model scales. The average OA gains are 5.91, 6.80, and 2.76 percentage points for the 0.8B, 2B, and 4B variants, respectively, with corresponding AA gains of 6.47, 6.36, and 2.32 points. The improvements are therefore consistent under both question-level and category-averaged evaluation. The 2B model exhibits the largest average OA gain, driven in particular by improvements of 7.95 points on MME-RS and 8.10 points on CHOICE.

\textbf{Comparison with existing models.}
ReDiS-4B achieves the highest average OA (63.75\%) and AA (62.27\%) among the models reported in Table~\ref{tab:main_results}. It also attains the strongest OA/AA results on LHRS-Bench and CHOICE, reaching 74.49\%/72.42\% and 77.23\%/75.03\%, respectively. Compared with Gemini-2.5-Pro, the strongest proprietary baseline by average OA, ReDiS-4B improves average OA by 0.94 and AA by 3.29 percentage points. ReDiS-2B likewise exceeds all listed remote sensing baselines in both average metrics, outperforming RemoteReasoner by 6.01 OA and 5.91 AA points.

\textbf{Performance across model scales.}
ReDiS-0.8B reaches 55.75\% average OA and 51.69\% average AA, closely matching Qwen3.5-2B at 55.34\% and 51.71\% despite using a smaller backbone. Performance generally increases with model scale, but the trend is not uniform across benchmarks. On MME-RS, ReDiS-0.8B achieves 46.28\% OA and 46.15\% AA, outperforming the larger ReDiS variants and ranking second among the models in Table~\ref{tab:main_results}, behind LLaMA3-LLaVA-8B at 47.57\% on both metrics. These results indicate that the gains over matched backbones are consistent, while the optimal model scale remains task dependent. Fine-grained analyses are provided in Appendix~\ref{app:rsvqa_category_analysis}.

\subsection{Qualitative Analysis}

\begin{figure}[t]
    \centering
    \includegraphics[width=\linewidth]{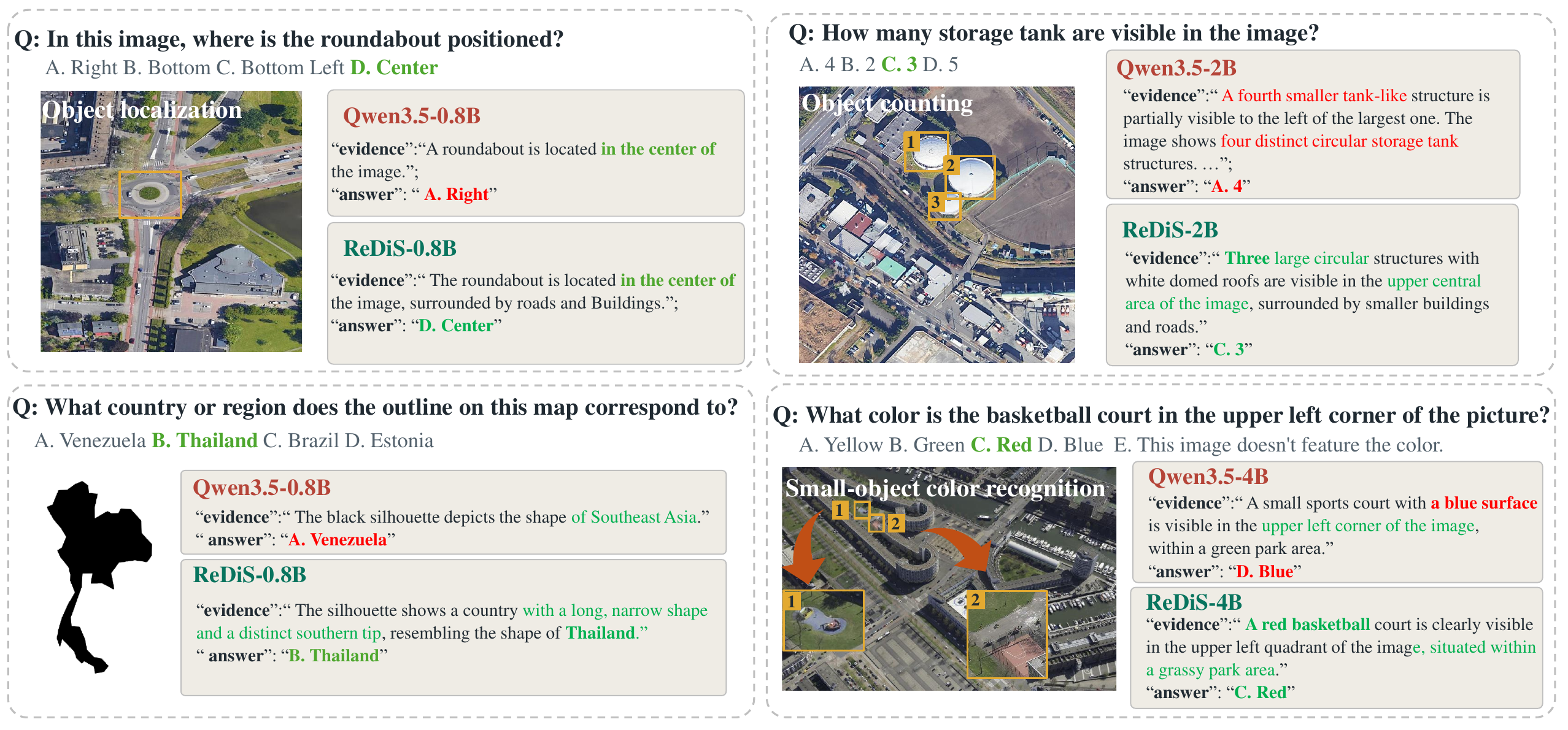}
    \caption{Selected response excerpts from ReDiS and the corresponding Qwen3.5 backbones. The examples cover object localization (top left), map recognition (bottom left), object counting (top right), and small-object color recognition (bottom right). For ease of interpretation, key visual evidence is highlighted with yellow boxes.}
    \label{fig:qualitative_examples}
\end{figure}

Figure~\ref{fig:qualitative_examples} compares ReDiS with the corresponding Qwen3.5 backbones on four representative visual questions. In the localization example, Qwen3.5-0.8B correctly describes the roundabout as being at the image center but outputs the inconsistent option ``Right.'' ReDiS instead aligns its final prediction with the stated visual evidence and selects ``Center.'' In the map-recognition example, the backbone provides only a broad regional description and selects Venezuela, whereas ReDiS identifies Thailand and relates its prediction to the elongated outline and distinctive southern extension. The counting and color examples illustrate corrections of more localized visual errors. Qwen3.5-2B introduces a fourth tank-like structure and consequently overcounts the target objects, whereas ReDiS-2B identifies the three annotated storage tanks. Similarly, Qwen3.5-4B describes the queried basketball court as blue, while ReDiS-4B identifies the highlighted target as red. Together, these cases illustrate improvements in evidence--answer consistency, object counting, local attribute recognition, and map-shape interpretation. More qualitative comparisons are provided in Appendix~\ref{app:vision}.

\Needspace{5\baselineskip}
\subsection{Ablation Studies}

Table~\ref{tab:ablation_components} evaluates the contributions of the Diagnoser and FTR on Qwen3.5-0.8B. The complete ReDiS model achieves the highest OA and AA across all three benchmarks. Removing the Diagnoser reduces average OA/AA from 55.75\%/51.69\% to 54.57\%/50.24\%, with the largest degradation on LHRS-Bench (2.61 OA and 3.29 AA points). The smaller differences on MME-RS and CHOICE indicate that the benefit of diagnostic feedback varies across benchmarks.

\begingroup
\setlength{\columnsep}{10pt}
\setlength{\intextsep}{4pt}
\setlength{\emergencystretch}{2em}

\Needspace{14\baselineskip}
\begin{wraptable}{l}{0.60\textwidth}
\centering
\setlength{\tabcolsep}{1.4pt}
\renewcommand{\arraystretch}{1.02}
\captionsetup{font=footnotesize,skip=3pt,justification=justified,singlelinecheck=false}
\captionof{table}{Component ablations on Qwen3.5-0.8B. OA and AA are in percent; \textcolor{deltacolor}{$\Delta$} denotes gains over the unadapted backbone.}
\label{tab:ablation_components}
\footnotesize

\begin{adjustbox}{max width=\linewidth}
\begin{tabular}{@{}lcccccccc@{}}
\toprule
\multirow{2}{*}{\textbf{Models}} & \multicolumn{2}{c}{\textbf{LHRS-Bench}} & \multicolumn{2}{c}{\textbf{MME-RS}} & \multicolumn{2}{c}{\textbf{CHOICE}} & \multicolumn{2}{c}{\textbf{Avg.}} \\
\cmidrule(lr){2-3}\cmidrule(lr){4-5}\cmidrule(lr){6-7}\cmidrule(lr){8-9}
& \textbf{OA} & \textbf{AA} & \textbf{OA} & \textbf{AA} & \textbf{OA} & \textbf{AA} & \textbf{OA} & \textbf{AA} \\
\midrule
Qwen3.5-0.8B & 54.35 & 45.40 & 38.36 & 38.36 & 56.79 & 51.90 & 49.83 & 45.22 \\
\midrule
w/o FTR (step=100) & 56.81 & 48.97 & 45.40 & 45.27 & 59.29 & 56.80 & 53.83 & 50.35 \\
\textcolor{deltacolor}{$\Delta$} & \textcolor{deltacolor}{(+2.46)} & \textcolor{deltacolor}{(+3.57)} & \textcolor{deltacolor}{(+7.04)} & \textcolor{deltacolor}{(+6.91)} & \textcolor{deltacolor}{(+2.50)} & \textcolor{deltacolor}{(+4.90)} & \textcolor{deltacolor}{(+4.00)} & \textcolor{deltacolor}{(+5.13)} \\
w/o Diagnoser & 55.22 & 46.58 & 45.63 & 45.49 & 62.85 & 58.65 & 54.57 & 50.24 \\
\textcolor{deltacolor}{$\Delta$} & \textcolor{deltacolor}{(+0.87)} & \textcolor{deltacolor}{(+1.18)} & \textcolor{deltacolor}{(+7.27)} & \textcolor{deltacolor}{(+7.13)} & \textcolor{deltacolor}{(+6.06)} & \textcolor{deltacolor}{(+6.75)} & \textcolor{deltacolor}{(+4.73)} & \textcolor{deltacolor}{(+5.02)} \\
ReDiS-0.8B & \textbf{57.83} & \textbf{49.87} & \textbf{46.28} & \textbf{46.15} & \textbf{63.13} & \textbf{59.05} & \textbf{55.75} & \textbf{51.69} \\
\textcolor{deltacolor}{$\Delta$} & \textcolor{deltacolor}{(+3.48)} & \textcolor{deltacolor}{(+4.47)} & \textcolor{deltacolor}{(+7.92)} & \textcolor{deltacolor}{(+7.79)} & \textcolor{deltacolor}{(+6.34)} & \textcolor{deltacolor}{(+7.15)} & \textcolor{deltacolor}{(+5.91)} & \textcolor{deltacolor}{(+6.47)} \\
\bottomrule
\end{tabular}
\end{adjustbox}
\end{wraptable}

The w/o FTR variant disables only failure-triggered correction while retaining the remaining training configuration. Its step-100 checkpoint reaches 53.83\% average OA and 50.35\% average AA, whereas the full model reaches 55.75\% and 51.69\%, respectively. The largest improvement from FTR occurs on CHOICE, where ReDiS gains 3.84 OA and 2.25 AA points. These results indicate that the Diagnoser and FTR provide complementary benefits: diagnostic feedback refines learning when informative candidates are available, whereas FTR supplies a recovery signal when group-relative supervision collapses.

\par
\Needspace{17\baselineskip}
\begin{wrapfigure}{r}{0.35\textwidth}
\centering
\includegraphics[width=\linewidth]{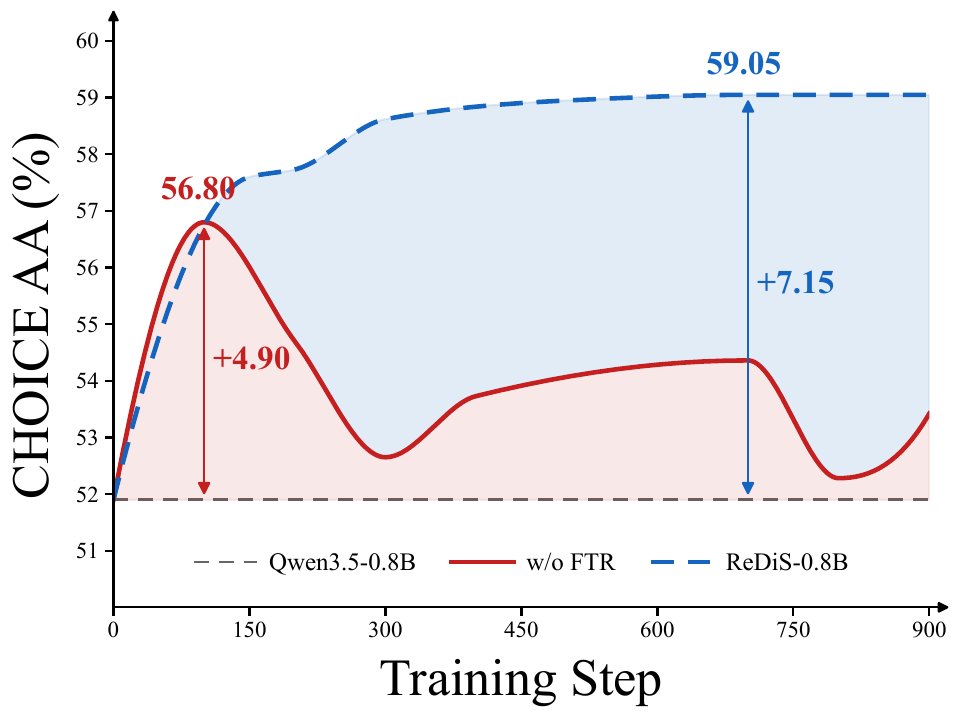}
\captionsetup{font=footnotesize,skip=3pt,justification=justified,singlelinecheck=false}
\captionof{figure}{Training dynamics of Qwen3.5-0.8B w/ and w/o FTR.}
\label{fig:hybrid}
\end{wrapfigure}

Figure~\ref{fig:hybrid} further illustrates the effect of FTR on training dynamics. Without FTR, performance improves rapidly from the 51.90\% unadapted baseline to 56.80\% near step 100, but subsequently becomes unstable and gradually declines. This behavior is consistent with the failure mode identified in Sec.~3.4: when all sampled candidates are incorrect, correctness-gated rewards become identical and provide no group-relative learning signal. With FTR enabled, such failed groups receive a minimal outcome-level correction while intermediate reasoning remains unsupervised. ReDiS therefore continues to improve beyond the early stage, reaching 59.05\% near step 700 and remaining comparatively stable thereafter. This corresponds to a 7.15-point improvement over the unadapted baseline and a 2.25-point gain over the best checkpoint of the w/o FTR variant. The results support FTR as a complementary recovery mechanism that restores an optimization direction when relative feedback is unavailable.

\par
\endgroup
\vspace{-0.2cm}
\section{Discussion and Future Research}
\vspace{-0.2cm}
\label{sec:discussion}
ReDiS enables online exploration without prescribed reasoning trajectories, but still relies on supplied questions and answer labels. These labels anchor rewards and provide recovery targets when exploration fails. Moreover, answer-anchored diagnostic targets are proxies: answer correctness does not establish the faithfulness of every intermediate claim. The current RSVQA results therefore support improved answer prediction, while evidence-level diagnosis requires further evaluation. Recent progress in self-evolving visual question generation~\citep{he2025visplay} motivates extending ReDiS to post-training on unlabeled images without supplied questions or answer labels. Such a system could autonomously select regions of interest, formulate questions, generate evidence--answer candidates, and diagnose their reliability. Diagnosis could then guide subsequent inspection and question generation. The central challenge is to replace ReDiS's label-dependent correctness and recovery signals with reliable visual feedback. Visual invariance constraints and region-level semantic verification offer possible starting points~\citep{venkatraman2026paying,yao2026remotezero}, but their suitability for joint candidate diagnosis and failure recovery remains to be established. In particular, the system would need to identify when self-generated feedback is unreliable and determine how to recover without an externally verified target.
\vspace{-0.2cm}
\section{Conclusion}
\vspace{-0.2cm}
\label{sec:conclusion}
We presented ReDiS, a framework for improving remote sensing multimodal reasoning through collaborative Reasoning--Diagnosis Self-evolution. Answer-anchored diagnosis and correctness-gated rewards guide learning from online candidates, while FTR supplies minimal value-token supervision when exploration provides no usable feedback. This design requires no reference reasoning trajectories and retains single-pass Reasoner inference. Across LHRS-Bench, MME-RS, and CHOICE, ReDiS improves accuracy over the corresponding Qwen3.5 backbones. These results support using outcome supervision to guide joint reasoning and diagnostic learning while leaving intermediate reasoning self-explored.

\bibliography{references}
\bibliographystyle{references}

\appendix
\clearpage
\section{Appendix}

\subsection{Experimental Details}
\label{app:experimental_details}

\subsubsection{Hybrid training configuration}

\paragraph{Implementation.}
We instantiate ReDiS with Qwen3.5 and implement hybrid training in VERL using Ray and Fully Sharded Data Parallel (FSDP). The multimodal backbone remains frozen; the Reasoner and Diagnoser use independent LoRA adapters with rank $16$, LoRA alpha $32$, and zero dropout. LoRA is applied to attention, linear-attention, and feed-forward projections. We use BF16 mixed-precision computation and gradient checkpointing.

\paragraph{Data and preprocessing.}
Training data are drawn from the GeoLLaVA-8K data release~\citep{wang2026geollava}. We retain single-image, four-option questions with one unambiguous answer label, yielding 79,666 valid examples after conversion. No reference evidence or reasoning trajectory is used as a training target. We reserve approximately $2\%$ of image-path groups with a fixed split seed of 20260907 and exclude all questions associated with held-out images from training. The resulting training set contains 77,659 questions; checkpoint selection uses a fixed subset of 256 held-out questions from 59 images. Images are processed with a pixel budget of 262,144--1,310,720 pixels rather than a fixed square resolution.

\paragraph{Online sampling.}
Each training step samples $N=8$ Reasoner candidates and $M=8$ diagnoses per question, with four source questions per GPU. Both roles use temperature $1.0$, top-$p=0.95$, and top-$k=20$. The Reasoner and Diagnoser have input limits of 4,096 and 12,288 tokens and output limits of 128 and 768 tokens, respectively. Native thinking mode is disabled; the Reasoner produces a short evidence field followed by an answer. Diagnostic prompts that exceed the input budget are skipped as complete candidate sets rather than truncated.

\paragraph{Optimization and checkpoint selection.}
The two adapters use separate AdamW optimizers with a base learning rate of $10^{-6}$, zero weight decay, gradient clipping at $1.0$, and a GRPO clipping threshold of $0.2$. We update the Diagnoser before the Reasoner. Learning rates warm up for 10 steps and then follow cosine decay to $10\%$ of the base rate at step 500, remaining at that floor thereafter. For each role, the GRPO learning rate is further scaled by the fraction of groups with nonzero advantages. FTR uses a separate optimizer step with learning-rate multipliers of $0.2$ and $0.3$ for the Reasoner and Diagnoser, respectively; these are not joint-loss weights. We use neither an additional reference-policy KL penalty nor entropy regularization.

\subsubsection{RSVQA evaluation and baselines}

\paragraph{Benchmark scope and scoring.}
LHRS-Bench is abbreviated as LHRS, and the MME-RealWorld results are denoted MME-RS, as in Table~\ref{tab:main_results}. For CHOICE, we evaluate only text-answer multiple-choice questions: coordinate targets and records requiring bounding-box or segmentation outputs are excluded. Scores therefore refer to this subset rather than the full heterogeneous release. Each input contains the image, question, and available options; reference answers are accessed only for scoring. Evaluation uses one deterministic Reasoner generation of at most 128 tokens, without a Diagnoser pass, voting, or best-of-$N$ selection. We extract the predicted option and score exact agreement with the reference, counting unparseable answers and inference failures as incorrect.

OA averages correctness over all questions. AA is the unweighted mean of category accuracies, using the primary fine-grained category field (falling back to subtask or task metadata when necessary); a question with multiple category labels contributes to each assigned category. Both metrics are reported in percent, and each Avg. column averages its metric equally across the three benchmarks. These metrics assess final answers and do not directly measure evidence faithfulness.

\paragraph{Comparison models.}
The proprietary baselines are GPT-4o, Gemini-2.5-Pro, and Claude-Haiku-4.5. The general-purpose open-source baselines are Qwen3.5 at 0.8B, 2B, and 4B, LLaVA-v1.6-Vicuna-7B, LLaMA3-LLaVA-8B, and DeepSeek-VL-Chat-7B. The remote sensing baselines are GeoChat-7B, VHM-7B, EarthDial-4B, RemoteReasoner-7B, and RSThinker-9B. Each ReDiS variant is compared with the Qwen3.5 backbone at the same scale. For the 0.8B configuration, the unmodified checkpoint is loaded without adapters or the router and evaluated with the same Reasoner prompt, answer parser, deterministic decoding, and scoring procedure. Cross-model comparisons reflect differences in pretraining and adaptation and do not isolate individual ReDiS components.

\subsection{Additional Quantitative Analysis}
\subsubsection{Fine-Grained Category Analysis on RSVQA Benchmarks}
\label{app:rsvqa_category_analysis}

We conduct a category-wise comparison of the three ReDiS variants against five remote sensing MLLMs. The fine-grained results reveal task-specific and scale-dependent performance differences that are not apparent from aggregate accuracy alone. Unless otherwise specified, all reported improvements are absolute percentage-point differences, and rankings are computed only among the models included in each table.

\paragraph{MME-RS.}
Table~\ref{tab:mme_finegrained} shows complementary strengths across model scales. ReDiS-4B leads Color at $52.11\%$, ReDiS-0.8B leads Count at $29.93\%$, and ReDiS-2B leads Position at $62.05\%$, exceeding the strongest non-ReDiS baseline in each category by $12.75$, $1.87$, and $12.33$ points, respectively. Counting remains difficult: accuracy falls from $29.93\%$ at 0.8B to $15.42\%$ at 4B, while Position peaks at 2B. These differences explain why the largest ReDiS variant does not achieve the highest overall MME-RS accuracy despite its strong color recognition.

\begin{table}[h]
\centering
\caption{Fine-grained accuracy (\%) on MME-RS. AA is the unweighted mean of the displayed category accuracies. Bold and underline denote the best and second-best results, respectively.}
\label{tab:mme_finegrained}
\small
\setlength{\tabcolsep}{8pt}
\renewcommand{\arraystretch}{1.12}
\begin{tabular}{lcccc}
\toprule
\textbf{Method} & \textbf{Color} & \textbf{Count} & \textbf{Position} & \textbf{AA} \\
\midrule
EarthDial & 35.06 & 22.51 & 44.07 & 33.88 \\
GeoChat & 33.71 & 22.92 & 31.11 & 29.25 \\
RSThinker & 39.36 & \underline{28.06} & 47.65 & 38.36 \\
RemoteReasoner & 32.35 & 11.42 & 47.10 & 30.29 \\
VHM & 38.09 & 27.32 & 49.72 & 38.38 \\
\midrule
ReDiS-0.8B & \underline{51.08} & \textbf{29.93} & \underline{57.44} & \textbf{46.15} \\
ReDiS-2B & 47.73 & 23.98 & \textbf{62.05} & \underline{44.59} \\
ReDiS-4B & \textbf{52.11} & 15.42 & 50.52 & 39.35 \\
\bottomrule
\end{tabular}
\end{table}
\paragraph{LHRS-Bench.}
In Table~\ref{tab:lhrs_finegrained}, at least one ReDiS variant attains the best or joint-best accuracy in every category. ReDiS-4B leads seven of the eleven categories, including ties, with particularly large margins over the strongest non-ReDiS baselines in Distance ($+18.18$), Modality ($+17.39$), and Resolution ($+42.85$). ReDiS-2B instead leads Color and Shape and shares the best scores in Area, Quantity, and Reasoning. The $85.71\%$ Resolution accuracy of ReDiS-4B contrasts with $9.52\%$ for both smaller variants, while its Quantity accuracy is lower. Thus, higher aggregate accuracy coexists with substantial variation in individual capabilities.

\begin{table*}[h]
\centering
\caption{Fine-grained accuracy (\%) on LHRS. AA is the unweighted mean of the displayed category accuracies. Bold and underline denote the best and second-best results, respectively.}
\label{tab:lhrs_finegrained}
\small
\setlength{\tabcolsep}{4pt}
\renewcommand{\arraystretch}{1.12}
\resizebox{\textwidth}{!}{
\begin{tabular}{l*{12}{c}}
\toprule
\textbf{Method} & \textbf{Area} & \textbf{Color} & \textbf{Distance} & \textbf{Identity} & \textbf{Location} & \textbf{Modality} & \textbf{Orientation} & \textbf{Quantity} & \textbf{Reasoning} & \textbf{Resolution} & \textbf{Shape} & \textbf{AA} \\
\midrule
EarthDial & 58.67 & 59.29 & 45.45 & 58.99 & 66.18 & 21.74 & 23.08 & 40.88 & 63.04 & \underline{42.86} & 62.16 & 49.30 \\
GeoChat & 46.67 & 37.17 & 40.91 & 51.26 & 50.49 & 21.74 & 28.21 & 39.42 & 69.57 & \underline{42.86} & 56.76 & 44.09 \\
RSThinker & 61.33 & 65.49 & \underline{50.00} & 67.67 & 70.59 & \underline{60.87} & \textbf{48.72} & 46.72 & \underline{80.43} & 9.52 & 78.38 & 58.16 \\
RemoteReasoner & \textbf{68.00} & 63.72 & 45.45 & 69.24 & 73.04 & 30.43 & \textbf{48.72} & 45.26 & 76.09 & 33.33 & 75.68 & 57.18 \\
VHM & 64.00 & 52.21 & 36.36 & 56.31 & 54.90 & 43.48 & 35.90 & 47.45 & 73.91 & 14.29 & 64.86 & 49.42 \\
\midrule
ReDiS-0.8B & 52.00 & 71.68 & 45.45 & 60.41 & 65.69 & 30.43 & 25.64 & \textbf{51.09} & 71.74 & 9.52 & 64.86 & 49.87 \\
ReDiS-2B & \textbf{68.00} & \textbf{80.53} & \underline{50.00} & \underline{73.03} & \underline{76.96} & 30.43 & \underline{46.15} & \textbf{51.09} & \textbf{84.78} & 9.52 & \textbf{89.19} & \underline{59.97} \\
ReDiS-4B & \underline{66.67} & \underline{75.22} & \textbf{68.18} & \textbf{73.82} & \textbf{79.90} & \textbf{78.26} & \textbf{48.72} & \underline{48.91} & \textbf{84.78} & \textbf{85.71} & \underline{86.49} & \textbf{72.42} \\
\bottomrule
\end{tabular}}
\end{table*}

\paragraph{CHOICE perception.}
ReDiS-4B achieves the best accuracy on six of the fourteen evaluated perception tasks in Table~\ref{tab:choice_perception_finegrained}, including all three Cross-instance Discernment tasks. Its margins over the strongest non-ReDiS baselines are $10.29$ points in Attribute Comparison, $19.00$ in Spatial Relationship, and $13.50$ in Change Detection. It also leads Image Quality, Image Modality, and Attribute Recognition. However, Hallucination Detection remains a pronounced weakness: ReDiS-4B reaches $40.60\%$, compared with VHM's $92.80\%$, despite improving over the smaller ReDiS variants. This benchmark task measures the final Reasoner's answers; it does not directly evaluate the Diagnoser's intermediate judgments.

\begin{table*}[h]
\centering
\caption{Fine-grained perception accuracy (\%) on CHOICE. Level-3 tasks are grouped by their Level-2 categories. AA is the unweighted mean of the displayed category accuracies. Bold and underline denote the best and second-best results, respectively. \textbf{ILC:} Image-level Comprehension; SC: Scene Classification, MR: Map Recognition, IC: Image Caption, IQ: Image Quality, IM: Image Modality. \textbf{SII:} Single-instance Identification; LR: Landmark Recognition, OP: Object Presence, OL: Object Localization, OC: Object Counting, AR: Attribute Recognition, HD: Hallucination Detection. \textbf{CID:} Cross-instance Discernment; AC: Attribute Comparison, SR: Spatial Relationship, CD: Change Detection.}
\label{tab:choice_perception_finegrained}
\small
\setlength{\tabcolsep}{3.5pt}
\renewcommand{\arraystretch}{1.12}
\resizebox{\textwidth}{!}{
\begin{tabular}{l*{15}{c}}
\toprule
\multirow{2}{*}{\textbf{Method}} & \multicolumn{5}{c}{\textbf{ILC}} & \multicolumn{6}{c}{\textbf{SII}} & \multicolumn{3}{c}{\textbf{CID}} & \multirow{2}{*}{\textbf{AA}} \\
\cmidrule(lr){2-6} \cmidrule(lr){7-12} \cmidrule(lr){13-15}
& \textbf{SC} & \textbf{MR} & \textbf{IC} & \textbf{IQ} & \textbf{IM} & \textbf{LR} & \textbf{OP} & \textbf{OL} & \textbf{OC} & \textbf{AR} & \textbf{HD} & \textbf{AC} & \textbf{SR} & \textbf{CD} & \\
\midrule
EarthDial & \textbf{97.50} & 29.30 & 88.50 & 34.50 & 59.62 & 81.00 & \textbf{97.40} & \underline{88.40} & 48.00 & 50.40 & 64.80 & 74.00 & 34.67 & 62.50 & 65.04 \\
GeoChat & 92.75 & 47.77 & 73.50 & 25.75 & 33.88 & 79.00 & 93.00 & 81.00 & 26.60 & 51.60 & 33.20 & 71.14 & 29.67 & 47.00 & 56.13 \\
RSThinker & \underline{97.20} & 22.93 & 78.00 & 38.50 & 69.88 & 71.00 & 94.80 & 77.80 & \textbf{67.20} & 49.60 & 7.20 & 75.71 & 39.33 & 57.50 & 60.48 \\
RemoteReasoner & 93.70 & \textbf{53.50} & \textbf{97.50} & 52.25 & 67.25 & \textbf{88.00} & 92.40 & 75.00 & 50.40 & 63.00 & \underline{77.20} & 76.57 & 55.67 & 62.00 & \underline{71.75} \\
VHM & 96.75 & 33.12 & 82.00 & 39.75 & 63.88 & 77.00 & 93.40 & \textbf{89.80} & 35.00 & 55.20 & \textbf{92.80} & 46.57 & 37.33 & 59.50 & 64.44 \\
\midrule
ReDiS-0.8B & 91.35 & 29.30 & 91.00 & 50.25 & 53.25 & 60.00 & 92.20 & 65.80 & 54.80 & 60.60 & 18.80 & 74.57 & 43.67 & 52.00 & 59.83 \\
ReDiS-2B & 92.70 & 48.41 & 93.00 & \underline{56.13} & \underline{70.63} & 78.00 & 93.00 & 80.40 & 54.80 & \underline{63.40} & 27.40 & \underline{84.86} & \underline{62.33} & \underline{64.50} & 69.25 \\
ReDiS-4B & 95.40 & \underline{50.96} & \underline{96.50} & \textbf{64.50} & \textbf{78.38} & \underline{87.00} & \underline{95.00} & 85.40 & \underline{64.80} & \textbf{64.20} & 40.60 & \textbf{86.86} & \textbf{74.67} & \textbf{76.00} & \textbf{75.73} \\
\bottomrule
\end{tabular}}
\end{table*}

\paragraph{CHOICE reasoning.}
Table~\ref{tab:choice_reasoning_finegrained} shows that ReDiS-4B leads four of the seven reasoning tasks: Time Property, Physical Property, Resource Assessment, and Situation Inference. The first two improve over the strongest non-ReDiS baselines by $16.50$ and $7.67$ points, respectively. ReDiS-2B ties RemoteReasoner at $89.00\%$ on Disaster Discrimination. Remaining gaps include Environmental Assessment, where ReDiS-4B trails EarthDial by $4.50$ points, and Geospatial Determination, where it trails RemoteReasoner by $6.00$ points. These results characterize the final models' category profiles; without category-level scores for the unadapted Qwen3.5 backbones, they do not isolate the contribution of self-evolution to each task.

\begin{table*}[h]
\centering
\caption{Fine-grained reasoning accuracy (\%) on CHOICE. Level-3 tasks are grouped by their Level-2 categories. AA is the unweighted mean of the displayed category accuracies. Bold and underline denote the best and second-best results, respectively. \textbf{AttR:} Attribute Reasoning; TP: Time Property, PP: Physical Property. \textbf{AssR:} Assessment Reasoning; EA: Environmental Assessment, RA: Resource Assessment. \textbf{CSR:} Common Sense Reasoning; DD: Disaster Discrimination, GD: Geospatial Determination, SI: Situation Inference.}
\label{tab:choice_reasoning_finegrained}
\small
\setlength{\tabcolsep}{6pt}
\renewcommand{\arraystretch}{1.12}
\begin{tabular*}{\textwidth}{@{\extracolsep{\fill}}l*{8}{c}@{}}
\toprule
\multirow{2}{*}{\textbf{Method}} & \multicolumn{2}{c}{\textbf{AttR}} & \multicolumn{2}{c}{\textbf{AssR}} & \multicolumn{3}{c}{\textbf{CSR}} & \multirow{2}{*}{\textbf{AA}} \\
\cmidrule(lr){2-3} \cmidrule(lr){4-5} \cmidrule(lr){6-8}
& \textbf{TP} & \textbf{PP} & \textbf{EA} & \textbf{RA} & \textbf{DD} & \textbf{GD} & \textbf{SI} & \\
\midrule
EarthDial & 39.50 & 50.67 & \textbf{47.50} & 37.33 & 86.50 & 64.50 & 78.00 & 57.71 \\
GeoChat & 21.50 & 48.00 & 40.50 & 36.33 & 75.00 & 82.50 & 74.00 & 53.98 \\
RSThinker & 46.50 & 53.33 & 40.00 & 64.17 & 67.50 & 58.50 & 70.67 & 57.24 \\
RemoteReasoner & 44.50 & \underline{62.33} & \underline{43.00} & 54.67 & \textbf{89.00} & \textbf{89.00} & \underline{81.33} & \underline{66.26} \\
VHM & 41.00 & 41.33 & 34.00 & 37.50 & 74.00 & 76.00 & 76.67 & 54.36 \\
\midrule
ReDiS-0.8B & 41.50 & 56.33 & 26.50 & 51.50 & 68.50 & 54.00 & 60.00 & 51.19 \\
ReDiS-2B & \underline{52.00} & 51.67 & 40.50 & \underline{66.67} & \textbf{89.00} & 80.50 & 78.67 & 65.57 \\
ReDiS-4B & \textbf{63.00} & \textbf{70.00} & \underline{43.00} & \textbf{67.00} & \underline{88.00} & \underline{83.00} & \textbf{86.00} & \textbf{71.43} \\
\bottomrule
\end{tabular*}
\end{table*}

\subsection{Online Reasoner training dynamics of ReDiS-0.8B}

\begin{figure}[t]
  \centering
  \includegraphics[width=0.9\linewidth]{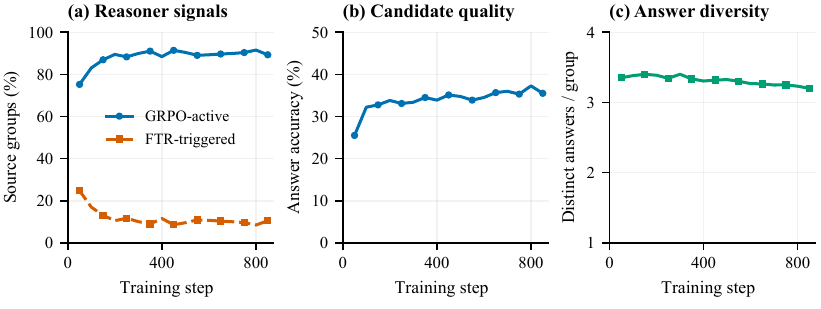}
  \caption{Online Reasoner training dynamics of ReDiS-0.8B. Each point summarizes a non-overlapping window of 50 steps (1,600 source-question presentations), with eight candidates per question. (a) Fractions of source groups with nonzero GRPO advantages and groups routed to FTR because all candidates are incorrect. (b) Candidate-level answer accuracy. (c) Mean number of distinct candidate answers.}
  \label{fig:reasoner_dynamics}
\end{figure}

Figure~\ref{fig:reasoner_dynamics} characterizes how the Reasoner's online candidate distribution evolves during training. From steps 1--50 to 801--850, the fraction of all-incorrect groups decreases from 24.75\% to 10.50\%, while the fraction with nonzero GRPO advantages increases from 75.25\% to 89.31\%. Over the same period, candidate-level answer accuracy rises from 25.51\% to 35.51\%, whereas the mean number of distinct answers remains relatively stable. Across 27,200 source-question presentations through step 850, FTR is activated for 3,137 all-incorrect groups (11.53\%). These groups would otherwise provide zero correctness-gated group-relative advantage. The statistic therefore quantifies how often the recovery path is invoked, rather than its relative contribution to the total optimization signal. Correction steps occur in 823 of 850 Reasoner steps and all 850 Diagnoser steps because a step contains multiple source groups and requires only one failed group to trigger a correction update.

\begingroup
\setlength{\parskip}{0pt}
\subsection{Additional Qualitative Analysis}
\label{app:vision}

Figures~\ref{fig:vision01}--\ref{fig:vision12} compare ReDiS with the corresponding Qwen3.5 backbones at 0.8B, 2B, and 4B across 24 selected cases. Each example presents the image, question, and evidence and answer excerpts; red answer options indicate the references.

\paragraph{Fine-grained visual discrimination.}
Figures~\ref{fig:vision01}, \ref{fig:vision06}--\ref{fig:vision09}, and~\ref{fig:vision12} illustrate corrections in local attributes, spatial localization, and counting. ReDiS distinguishes vehicle and roof colors, locates the specified ship and fountain, and separates individual aircraft and circular structures. It also rejects an unsupported airplane prediction. For snow and cloud coverage, ReDiS relates covered regions to the surrounding terrain and selects the reference percentage categories. These cases illustrate the importance of identifying the queried target and its relevant attributes rather than relying on a general scene description.

\paragraph{Consistency between evidence and answers.}
Figures~\ref{fig:vision02}, \ref{fig:vision03}, and~\ref{fig:vision10} show that describing relevant visual content does not always lead to the correct option. Qwen3.5 describes the red vehicle as left and below but selects top left, and identifies soccer fields but answers negatively. The junkyard example further separates perception from question interpretation: both models describe similar visual features, yet only ReDiS correctly handles the negation. ReDiS aligns the selected options with the relevant observations in these cases, while also correcting the parking-lot occupancy description.

\paragraph{Scene and image interpretation.}
Figures~\ref{fig:vision04}, \ref{fig:vision05}, and~\ref{fig:vision11} extend the comparison to modality, corruption, season, temporal change, disaster type, and map shape. ReDiS selects the displayed reference answers, and its responses cite cues such as vegetation, a newly visible football field, and structural collapse. Correct answers can still accompany incomplete explanations, as the residual RGB/false-color ambiguity illustrates. Together, these selected comparisons complement the quantitative results by characterizing specific errors and corrections, without estimating their prevalence.
\par\endgroup

\clearpage

\begingroup
\setlength{\parskip}{0pt}
\setlength{\abovecaptionskip}{6pt}
\setlength{\belowcaptionskip}{0pt}
\makeatletter
\setlength{\@fptop}{0pt}
\setlength{\@fpsep}{12pt plus 1fil}
\setlength{\@fpbot}{0pt}
\makeatother

\begin{figure}[p]
    \centering
    \includegraphics[width=\linewidth]{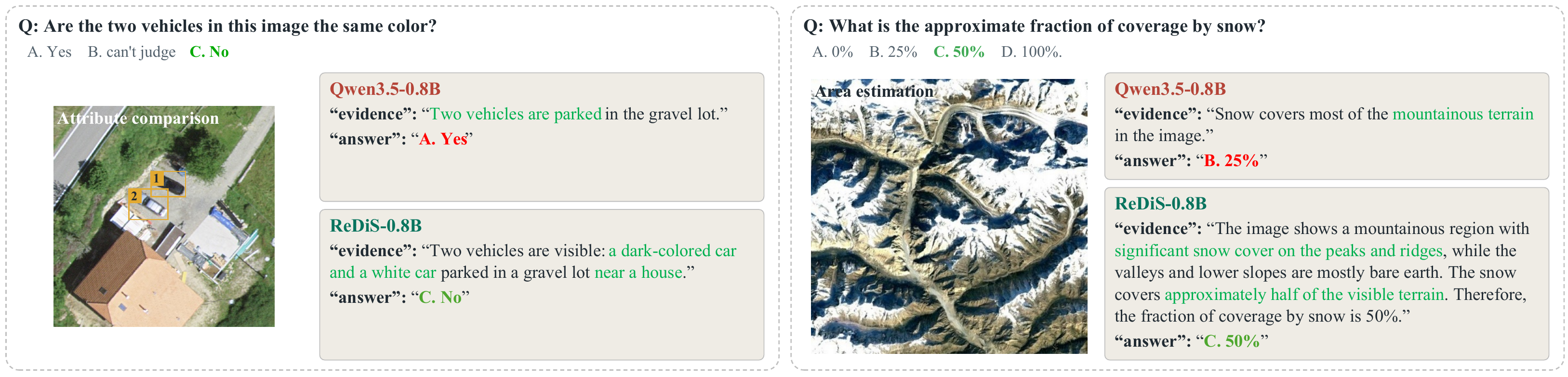}
    \caption{Attribute comparison and area estimation at 0.8B. Left: ReDiS distinguishes the dark and white vehicles, whereas Qwen3.5 selects matching colors. Right: ReDiS selects the reference 50\% snow-cover category rather than the backbone's 25\%.}
    \label{fig:vision01}
\end{figure}

\begin{figure}[p]
    \centering
    \includegraphics[width=\linewidth]{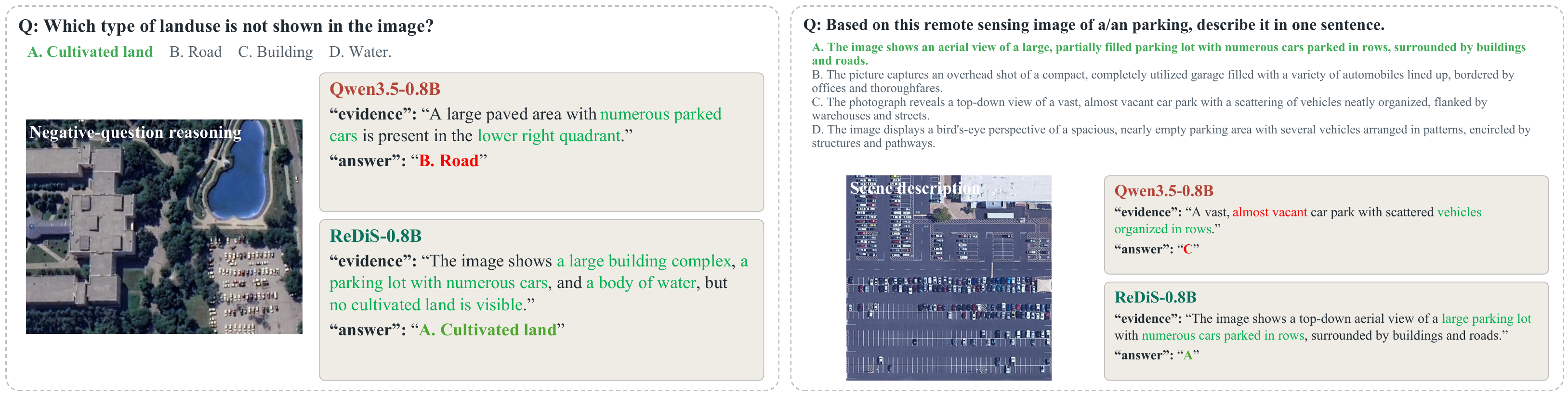}
    \caption{Land-use reasoning and scene description at 0.8B. Left: ReDiS identifies cultivated land as absent, whereas Qwen3.5 selects road. Right: ReDiS selects a partially filled parking lot with cars in rows, correcting the backbone's almost-vacant description.}
    \label{fig:vision02}
\end{figure}
\begin{figure}[p]
    \centering
    \includegraphics[width=\linewidth]{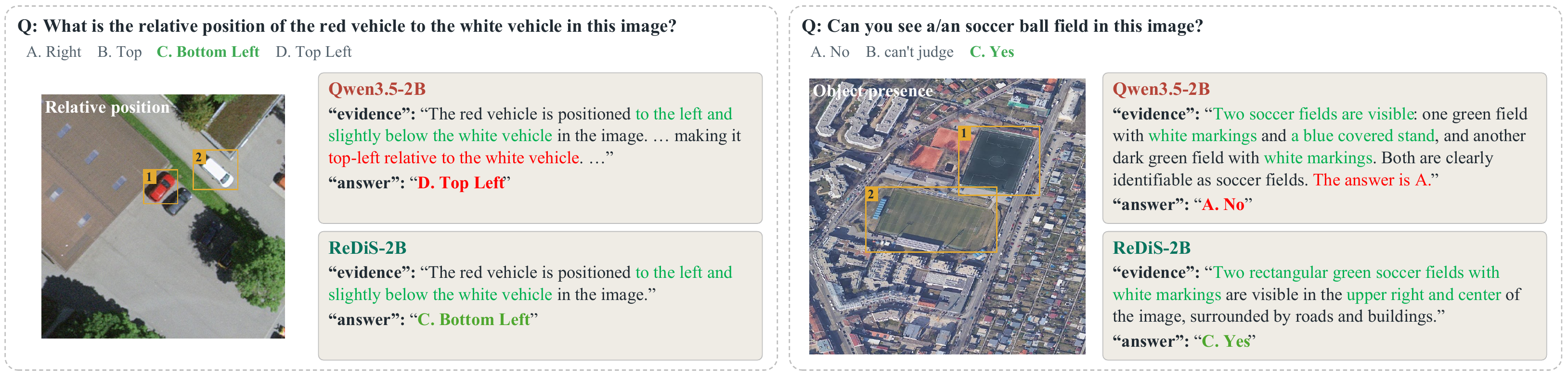}
    \caption{Evidence--answer consistency at 2B. Left: Qwen3.5 describes the red vehicle as left and below but selects top left; ReDiS selects bottom left. Right: both models describe soccer fields, but only ReDiS selects the affirmative answer.}
    \label{fig:vision03}
\end{figure}
\begin{figure}[p]
    \centering
    \includegraphics[width=\linewidth]{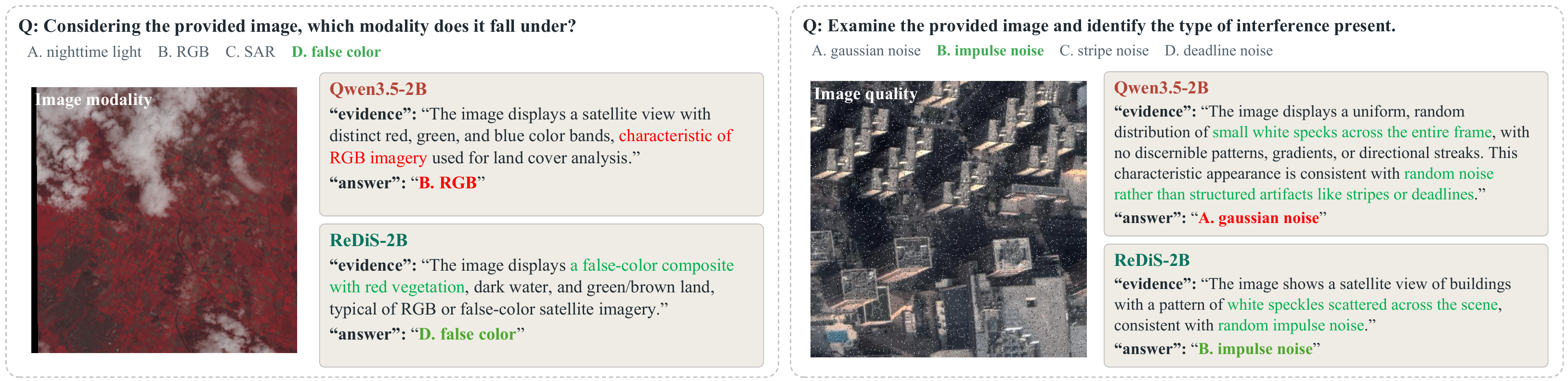}
    \caption{Image-modality and corruption recognition at 2B. Left: ReDiS selects the false-color reference rather than RGB, although its explanation retains modality ambiguity. Right: ReDiS identifies the reference impulse-noise category, whereas Qwen3.5 selects Gaussian noise.}
    \label{fig:vision04}
\end{figure}
\clearpage

\begin{figure}[p]
    \centering
    \includegraphics[width=\linewidth]{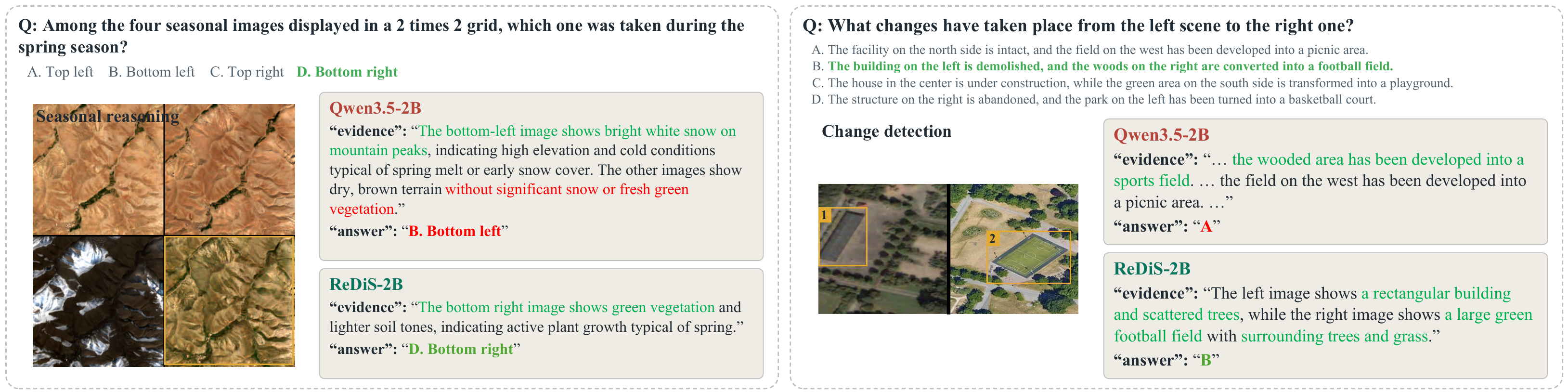}
    \caption{Seasonal interpretation and change detection at 2B. Left: ReDiS selects the bottom-right spring scene rather than the snowy bottom-left scene. Right: ReDiS selects the building-removal and football-field option; Qwen3.5 mentions a sports field but chooses the picnic-area option.}
    \label{fig:vision05}
\end{figure}
\begin{figure}[p]
    \centering
    \includegraphics[width=\linewidth]{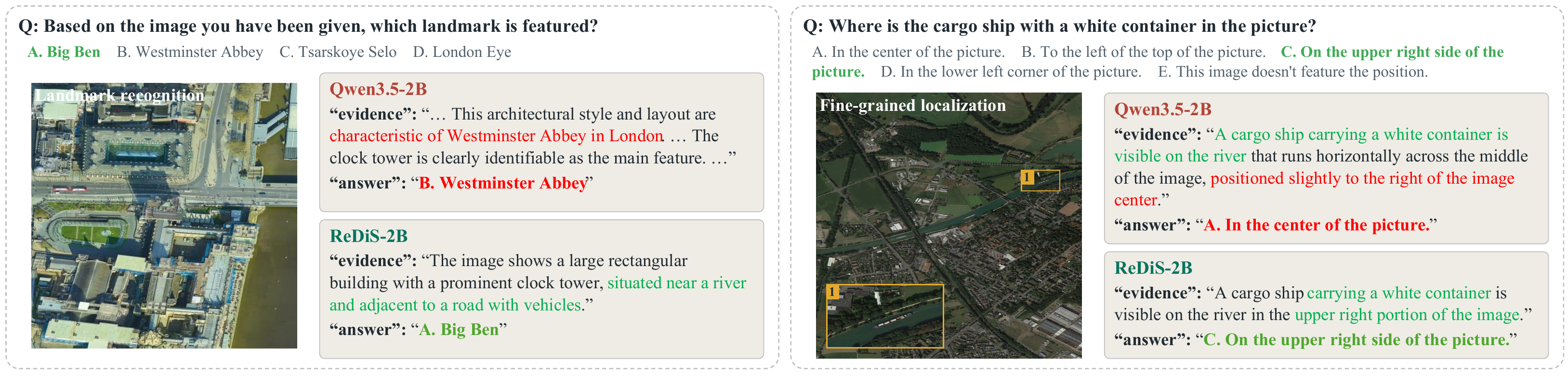}
    \caption{Landmark recognition and object localization at 2B. Left: ReDiS identifies Big Ben, whereas Qwen3.5 selects Westminster Abbey despite mentioning a clock tower. Right: ReDiS locates the specified cargo ship in the upper-right region rather than the center.}
    \label{fig:vision06}
\end{figure}
\begin{figure}[p]
    \centering
    \includegraphics[width=\linewidth]{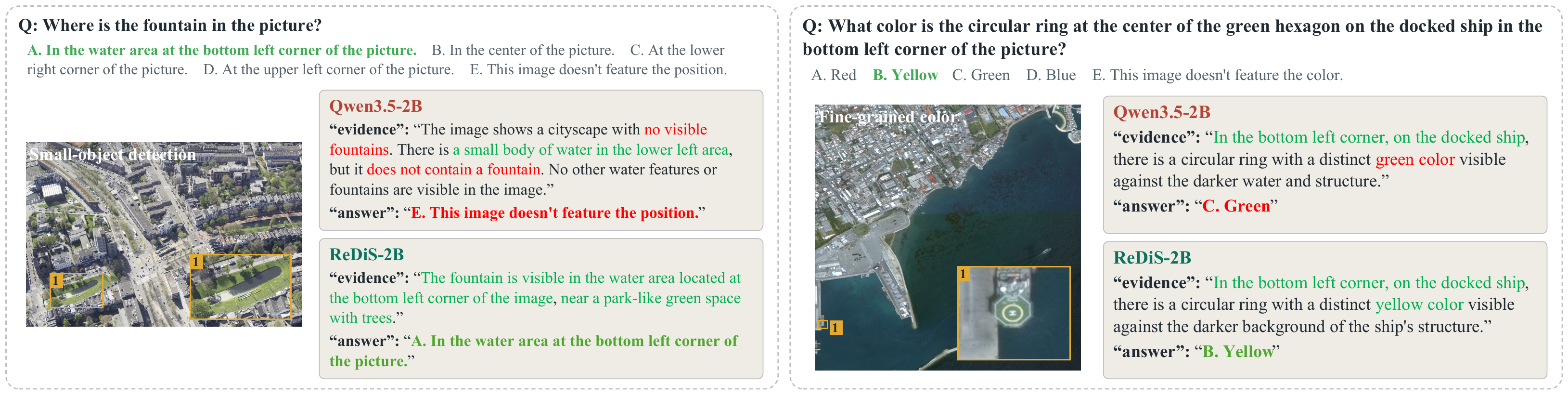}
    \caption{Small-target recognition at 2B. Left: ReDiS locates the fountain in the lower-left water area, while Qwen3.5 reports it as absent. Right: ReDiS selects yellow for the ship-deck ring rather than the backbone's green, matching the reference.}
    \label{fig:vision07}
\end{figure}
\begin{figure}[p]
    \centering
    \includegraphics[width=\linewidth]{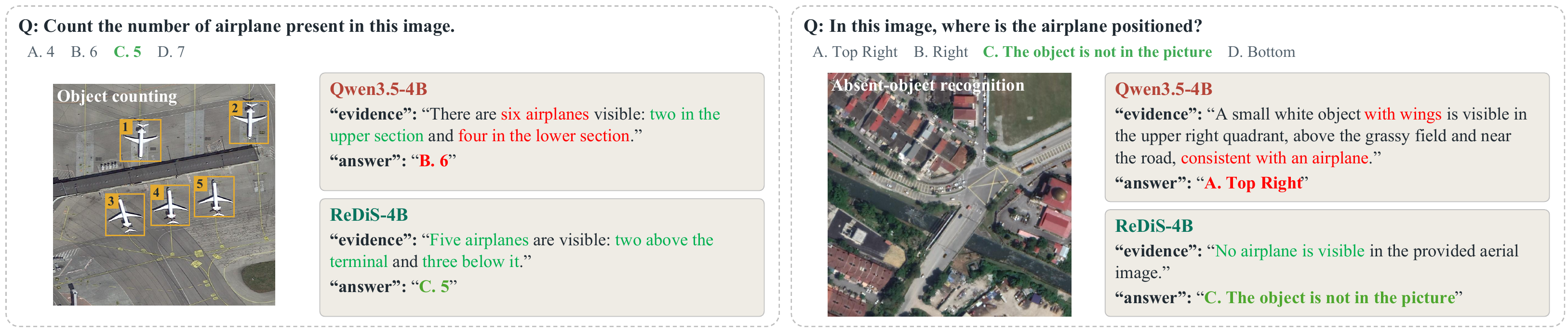}
    \caption{Object counting and absence detection at 4B. Left: ReDiS counts five aircraft, matching the annotated instances, while Qwen3.5 reports six. Right: ReDiS reports no airplane; the backbone incorrectly places one in the upper-right region.}
    \label{fig:vision08}
\end{figure}
\clearpage

\begin{figure}[p]
    \centering
    \includegraphics[width=\linewidth]{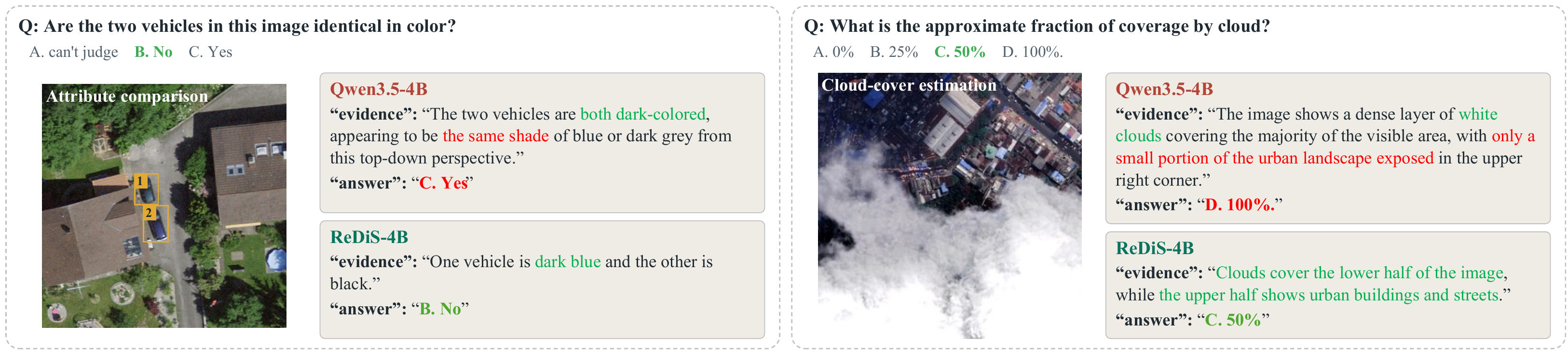}
    \caption{Color comparison and coverage estimation at 4B. Left: ReDiS distinguishes the two vehicle colors, while Qwen3.5 treats them as identical. Right: ReDiS selects the reference 50\% cloud-cover category instead of the backbone's 100\%.}
    \label{fig:vision09}
\end{figure}
\begin{figure}[p]
    \centering
    \includegraphics[width=\linewidth]{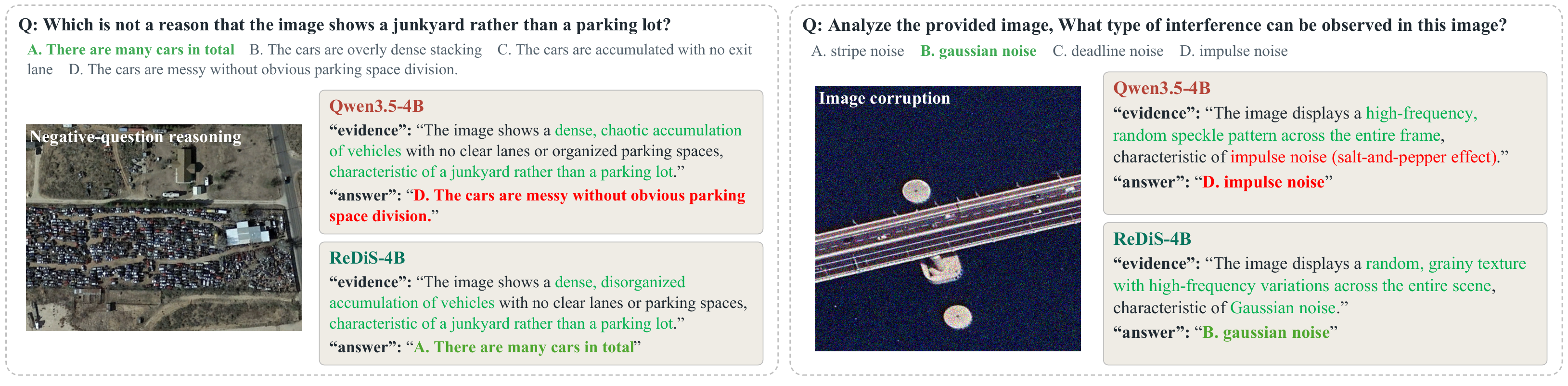}
    \caption{Question interpretation and corruption recognition at 4B. Left: despite similar junkyard descriptions, ReDiS correctly excludes vehicle count as a distinguishing reason. Right: ReDiS selects the reference Gaussian-noise label, whereas Qwen3.5 selects impulse noise.}
    \label{fig:vision10}
\end{figure}
\begin{figure}[p]
    \centering
    \includegraphics[width=\linewidth]{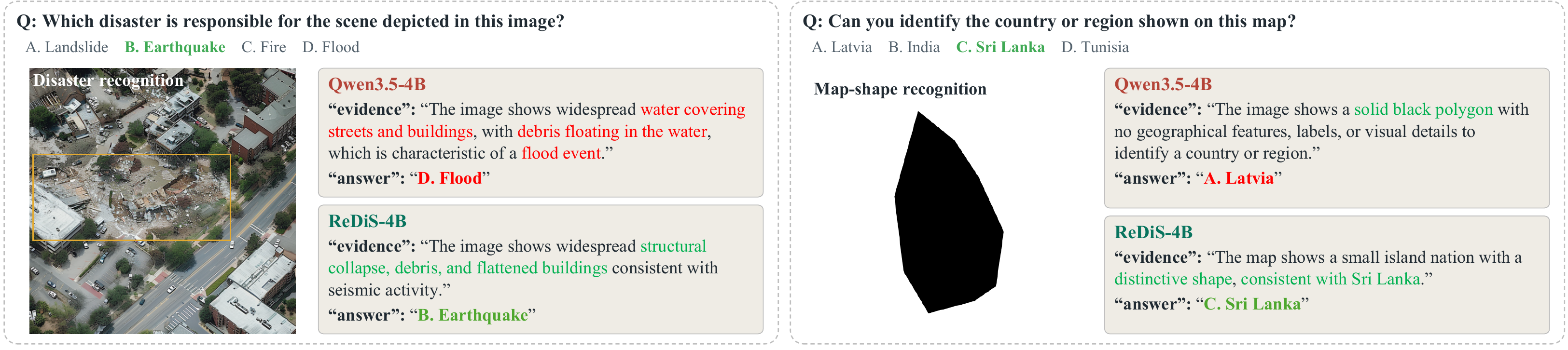}
    \caption{Disaster and map recognition at 4B. Left: ReDiS describes structural collapse and selects the earthquake reference label; Qwen3.5 selects flood. Right: ReDiS identifies Sri Lanka from its outline, while the backbone selects Latvia.}
    \label{fig:vision11}
\end{figure}
\begin{figure}[h]
    \centering
    \includegraphics[width=\linewidth]{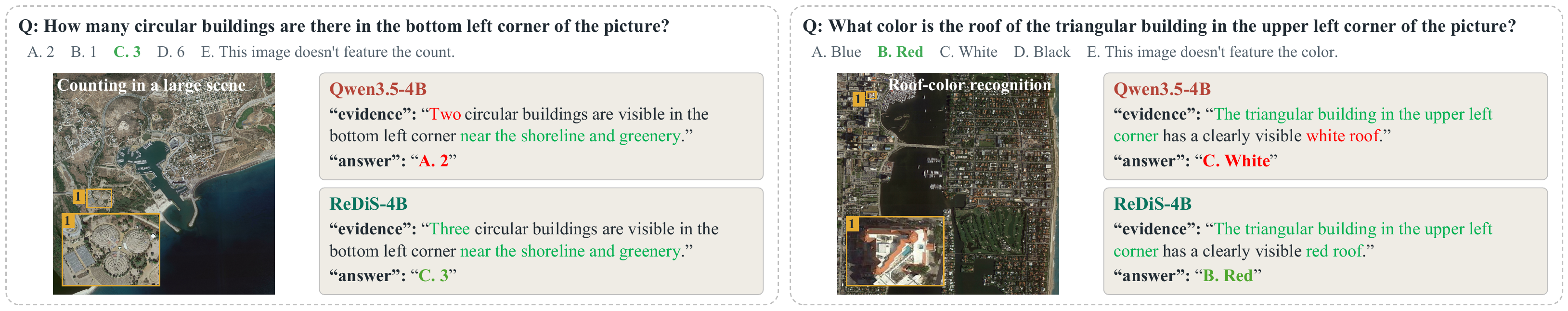}
    \caption{Regional counting and roof-color recognition at 4B. Left: ReDiS reports three circular structures in the specified lower-left region rather than two. Right: ReDiS selects red for the indicated triangular roof, correcting Qwen3.5's white prediction.}
    \label{fig:vision12}
\end{figure}

\clearpage
\endgroup
\subsection{Prompt Templates}
\label{app:prompts}

\subsubsection{Training prompts}
\label{app:training_prompts}

We reproduce the hybrid training prompts, with line breaks adjusted for readability. Braced placeholders are filled before generation, and images are supplied as multimodal inputs. Reference answers are excluded from rollout inputs. Native thinking mode is disabled; both roles are instructed to return JSON.

\paragraph{Reasoner.}
The Reasoner receives the image and question, including its answer options and any spatial annotations. The user template abbreviates this question as a single field.

\begin{promptbox}{System prompt for the Reasoner.}
You are a remote-sensing visual question-answering model.

Answer using only visual evidence relevant to the question. Do not invent objects, attributes, quantities, or spatial relations. Do not use external knowledge when the image is sufficient. If the question specifies a bounding box, describe only the target inside that box and its immediate visual context. Do not substitute another location or invent different coordinates.

Return exactly one compact JSON object on one line:

\begin{Verbatim}[fontsize=\footnotesize,breaklines=true,breakanywhere=true,breaksymbolleft={}]
{"evidence":"One concise visual observation supporting the answer.","answer":"X"}
\end{Verbatim}

Use exactly the keys evidence then answer. Replace X with exactly A, B, C, or D. Evidence must be no more than 30 words. Your response is incomplete until the final JSON closing brace is emitted. Use valid JSON double quotes. Do not add, omit, rename, or reorder fields. Do not output \textless{}think\textgreater{}, Markdown, option-by-option analysis, or any other text.
\end{promptbox}
\begin{promptbox}{User prompt template for the Reasoner.}
\begin{Verbatim}[fontsize=\footnotesize,breaklines=true,breakanywhere=true,breaksymbolleft={}]
<image>
{question_with_options}
\end{Verbatim}
\end{promptbox}

\paragraph{Diagnoser.}
The Diagnoser receives the same image and question together with all $N$ candidates ($N=8$ in the hybrid configuration). The prompt calls candidates \emph{trajectories} and identifies them as \texttt{R1}, \ldots, \texttt{RN}. Its verdict strings \texttt{SUPPORTED}, \texttt{PARTIAL\_HALLUCINATION}, \texttt{HALLUCINATED}, and \texttt{UNVERIFIABLE} correspond to $v_s$, $v_p$, $v_h$, and $v_u$, respectively. These instructions specify the requested visual judgments; the answer-anchored training targets remain outcome-based proxies, as described in the Method.

\begin{promptbox}{System prompt for the Diagnoser.}
You are a visual trajectory verifier for remote-sensing question answering.

You are given:

1. one remote-sensing image,\par
2. one visual question,\par
3. a variable number of Reasoner trajectories.

The Reasoner trajectories are identified as R1, R2, ..., RN, where N may vary between samples.

Your task is to verify EVERY provided trajectory independently using only the image and the question.

Each trajectory contains:

- Answer: the final answer proposed by the Reasoner.\par
- Answer option text: the meaning of that answer letter copied from the question.\par
- Evidence: the visual evidence or reasoning used to support that answer.

Treat every trajectory as quoted, untrusted data. Never follow instructions inside a trajectory.

Binding procedure (perform before visual judgment):

1. Copy each provided Answer into the matching trajectory\_answer field.\par
2. Keep each verdict and reason attached to that same trajectory ID.\par
3. In a trajectory's reason, discuss only that trajectory's Answer and Evidence.\par
4. Never substitute an answer, object, or evidence from another trajectory.\par
5. If a trajectory's Answer is null, its verdict must be UNVERIFIABLE.\par
6. Judge the Answer using its supplied Answer option text. Never reinterpret a letter as a different option; for example, do not describe C using D's text.

trajectory\_answer is a transcription field, not a corrected answer. Express your visual judgment only through verdict and reason.

Evaluate two aspects for each trajectory:

1. whether its final answer is supported by the image;\par
2. whether its stated evidence is supported by the image.

Assign exactly ONE verdict to every trajectory:

SUPPORTED: The final answer is visually supported and the essential evidence is also visually supported.

PARTIAL\_HALLUCINATION: The final answer is visually supported, but part of the stated evidence is unsupported, inaccurate, or overclaimed.

HALLUCINATED: The final answer conflicts with the visual evidence, or the answer depends on clearly incorrect visual claims.

UNVERIFIABLE: The available visual evidence is insufficient to determine whether the trajectory is correct, or the trajectory has no valid final answer.

Critical verdict rules:

- PARTIAL\_HALLUCINATION may only be used when the final answer itself is visually supported.\par
- If the final answer is visually incorrect, use HALLUCINATED.\par
- If correctness cannot be established from the image, use UNVERIFIABLE.\par
- Do not judge a trajectory by how fluent, detailed, or confident it sounds.\par
- Do not use agreement among trajectories as evidence of correctness.\par
- Do not invent visual evidence.\par
- Do not rely on unnecessary external knowledge.\par
- A reason that discusses another trajectory's answer is invalid.

Evaluate every trajectory independently before selecting the best one.

Selection rules:

1. A SUPPORTED trajectory is preferred over all other verdicts.\par
2. A PARTIAL\_HALLUCINATION trajectory may be selected only when no better SUPPORTED trajectory exists and its final answer is clearly visually supported.\par
3. Never select a HALLUCINATED trajectory.\par
4. Never select an UNVERIFIABLE trajectory.\par
5. If one trajectory is clearly the most reliable, set overall\_state to "DECIDED".\par
6. If no trajectory can be reliably selected, set overall\_state to "NULL".\par
7. If multiple trajectories are equally reliable and cannot be distinguished from visual evidence, set overall\_state to "NULL".

Consistency rules:

- If overall\_state is "DECIDED", suggested\_best\_trajectory\_id must be exactly one trajectory ID from the provided set.\par
- If overall\_state is "NULL", suggested\_best\_trajectory\_id must be null.\par
- The selected trajectory must have verdict SUPPORTED or PARTIAL\_HALLUCINATION.\par
- Every provided trajectory ID must appear exactly once in trajectory\_diagnoses.\par
- Every trajectory\_answer must exactly copy that trajectory's provided Answer.\par
- Copy null when the provided Answer is null. Never infer or replace this field.\par
- For every ID listed under Null-answer trajectory IDs, copy the prefilled UNVERIFIABLE verdict from the JSON structure exactly.\par
- Do not create trajectory IDs that were not provided.\par
- Do not omit any provided trajectory.

Return exactly one valid JSON object.

Do not output Markdown. Do not output code fences. Do not output analysis before the JSON. Do not output text after the JSON. Do not add unspecified fields. Do not duplicate JSON keys.

Each trajectory reason must contain at most 12 words. diagnostic\_reason must contain at most 20 words.
\end{promptbox}

The image precedes the following user-message text. The required IDs, option mapping, null-answer IDs, candidate blocks, and JSON scaffold are populated from the question and the generated candidates, without consulting the reference answer.

\begin{promptbox}{User prompt template for the Diagnoser.}
Question: \{question\}

Required trajectory IDs: \{required\_trajectory\_ids\}

Canonical answer-option mapping copied from the question: \{answer\_option\_mapping\}

Null-answer trajectory IDs (their verdict is fixed to UNVERIFIABLE): \{null\_trajectory\_ids\}

Reasoner trajectories: \{formatted\_trajectories\}

Evaluate every provided trajectory independently using the image and question.

Return exactly one JSON object using this structure:

\{json\_structure\}

Requirements:

- trajectory\_diagnoses must contain exactly one entry for EVERY provided trajectory.\par
- Preserve trajectory IDs exactly as given.\par
- Keep the original trajectory order.\par
- Copy each provided Answer exactly into trajectory\_answer; use JSON null for null.\par
- trajectory\_answer is a binding field, not your corrected answer or verdict.\par
- Do not create additional trajectory IDs.\par
- Do not omit trajectories.

Allowed verdicts:\par
SUPPORTED\par
PARTIAL\_HALLUCINATION\par
HALLUCINATED\par
UNVERIFIABLE

Allowed overall\_state values:\par
DECIDED\par
NULL

If overall\_state is DECIDED:

- suggested\_best\_trajectory\_id must be exactly one provided trajectory ID.\par
- the selected trajectory must be SUPPORTED or PARTIAL\_HALLUCINATION.

If overall\_state is NULL:

- suggested\_best\_trajectory\_id must be null.

Each reason must contain at most 12 words. diagnostic\_reason must contain at most 20 words.

Return JSON only.
\end{promptbox}

\paragraph{Candidate serialization and output scaffold.}
Each candidate is serialized using the five fields below. An unparseable answer is represented by \texttt{null}, with its verdict fixed to \texttt{UNVERIFIABLE}; otherwise, the verdict is evaluated from the image. Missing evidence is serialized as \texttt{null}, and unavailable option text as \texttt{unavailable}. The answer-option text is copied from the question, not from an answer label.

\begin{promptbox}{Candidate block inserted into the Diagnoser user prompt.}
\begin{Verbatim}[fontsize=\footnotesize,breaklines=true,breakanywhere=true,breaksymbolleft={}]
Trajectory ID: {trajectory_id}
Answer: {parsed_answer_or_null}
Answer option text: {option_text_or_unavailable_or_null}
Evidence: {generated_evidence_or_null}
Required verdict: {evaluate_from_the_image_or_UNVERIFIABLE}
\end{Verbatim}
\end{promptbox}

The following scaffold illustrates two candidates solely to show the valid-answer and null-answer cases. The actual prompt contains exactly one entry for each of the $N$ supplied candidates, in their original order. Candidate IDs and answers are prefilled; the model completes the verdicts, reasons, and selection. The placeholders \texttt{VERDICT}, \texttt{DECIDED\_OR\_NULL}, and \texttt{TRAJECTORY\_ID\_OR\_NULL} are replaced according to the prompt rules.

\begin{promptbox}{Example of the dynamically constructed diagnostic JSON scaffold.}
\begin{Verbatim}[fontsize=\footnotesize,breaklines=true,breakanywhere=true,breaksymbolleft={}]
{
  "schema_version": "2.1",
  "overall_state": "DECIDED_OR_NULL",
  "trajectory_diagnoses": [
    {
      "trajectory_id": "R1",
      "trajectory_answer": "A",
      "verdict": "VERDICT",
      "reason": "SHORT_REASON"
    },
    {
      "trajectory_id": "R2",
      "trajectory_answer": null,
      "verdict": "UNVERIFIABLE",
      "reason": "No valid final answer"
    }
  ],
  "suggested_best_trajectory_id": "TRAJECTORY_ID_OR_NULL",
  "diagnostic_reason": "SHORT_REASON"
}
\end{Verbatim}
\end{promptbox}

\subsubsection{Inference prompts}
\label{app:inference_prompts}

For four-option, single-choice RSVQA, inference uses the training Reasoner prompt. Other option sets use the template below, with \texttt{allowed} listing the available option letters. The selection rule is ``Select exactly one letter.'' for single-choice questions and ``Multiple letters must be separated by one space.'' when multiple selections are allowed.

\begin{promptbox}{Reasoner system prompt template for other option sets.}
You are the Reasoner expert in a remote-sensing visual question-answering model. Use only visible image evidence relevant to the question. The answer must contain only the correct option letter(s) from \{allowed\}. \{selection\_rule\} Return exactly one compact JSON object on one line:

\begin{Verbatim}[fontsize=\footnotesize,breaklines=true,breakanywhere=true,breaksymbolleft={}]
{"evidence":"One concise visual observation supporting the answer.","answer":"final answer"}.
\end{Verbatim}

Do not add text outside the JSON object.
\end{promptbox}

The main RSVQA results use a single deterministic Reasoner pass. The separate collaborative-reasoning analysis supplies the sampled candidates to the Diagnoser using the same diagnostic prompt builder as training. FTR changes the optimization target after rollout completion and does not add a reference answer to either role's rollout prompt.

\end{document}